\documentclass{iaii_pai}

\usepackage{amsmath,amsfonts}
\usepackage{algorithm}
\usepackage{algpseudocode}
\usepackage{enumitem}
\usepackage{tikz}
\usepackage{adjustbox}
\usepackage[percent]{overpic}

\newcommand{\figref}[1]{\cref{#1}}
\newcommand{\tabref}[1]{\cref{#1}}
\newcommand{\eqnref}[1]{\cref{#1}}
\newcommand{\secref}[1]{\cref{#1}}
\newcommand{\cmark}{\checkmark}
\newcommand{\xmark}{\texttimes}

\newcommand{\methodname}{PAIWorld}

\newcommand{\lossrepa}{\mathcal{L}_{\text{REPA}}}
\newcommand{\lossdiff}{\mathcal{L}_{\text{diff}}}
\newcommand{\losstotal}{\mathcal{L}_{\text{total}}}
\newcommand{\R}{\mathbb{R}}
\newcommand{\E}{\mathbb{E}}
\newcommand{\feat}{\mathbf{z}}

\graphicspath{{figures/}}

\title{PAIWorld: A 3D-Consistent World Foundation Model\\ for Robotic Manipulation}

\author{The PAIWorld Team}

\affiliation{Institute of AI for Industries, Chinese Academy of Sciences}

\date{June 2026}

\abstract{%
World foundation models (WFMs) have emerged as powerful simulators of physical environments, yet they predominantly operate in a single-view setting and lack the multi-view 3D consistency that robotic manipulation demands. Robotic systems inherently rely on multiple cameras, including egocentric, eye-to-hand, and wrist-mounted views, to capture complementary viewpoints for policy learning. Current multi-view world models, however, simply concatenate view tokens without explicit geometric reasoning, yielding cross-view object drift, depth inconsistency, and texture misalignment that propagate errors into downstream planning and control. We trace these failures to two fundamental deficiencies: (1)~the absence of an explicit \emph{inter-view communication mechanism}, which forces each viewpoint to generate in isolation, and (2)~the absence of a \emph{3D geometric prior}, which leaves the model without guidance on what constitutes physically correct cross-view structure. We argue that resolving both is necessary and sufficient: an information pathway across views and a geometric signal that steers it must coexist, since communication without geometric guidance collapses to trivial shortcuts, while geometric priors without an inter-view pathway cannot propagate constraints across viewpoints. Building on this analysis, we present \textbf{\methodname{}}, a framework that augments diffusion-transformer world models along two technical pillars, realized by three components. To build the \emph{inter-view communication pathway}, \textbf{Geometry-Aware Cross-View Attention} blocks open an explicit pathway across views, while \textbf{Geometric Rotary Position Embedding} encodes camera ray directions and extrinsic poses into this attention via rotary position encoding. To supply the \emph{geometric learning signal}, \textbf{Latent 3D-REPA} distills 3D-aware features from frozen 3D foundation models, ensuring the exchanged content is 3D-consistent. Built upon the DiT-based world foundation model, \methodname{} attains state-of-the-art multi-view 3D consistency on robotic manipulation benchmarks, ranking 1st on the WorldArena leaderboard and 2nd on the AgiBot-Challenge2026 leaderboard, and enables downstream applications including model-based planning, world action models, and multi-view policy post-training.%
}

\begin{document}

\maketitle

\section{Introduction}
\label{sec:introduction}

\begin{figure*}[t]
	\centering
	\includegraphics[width=\textwidth]{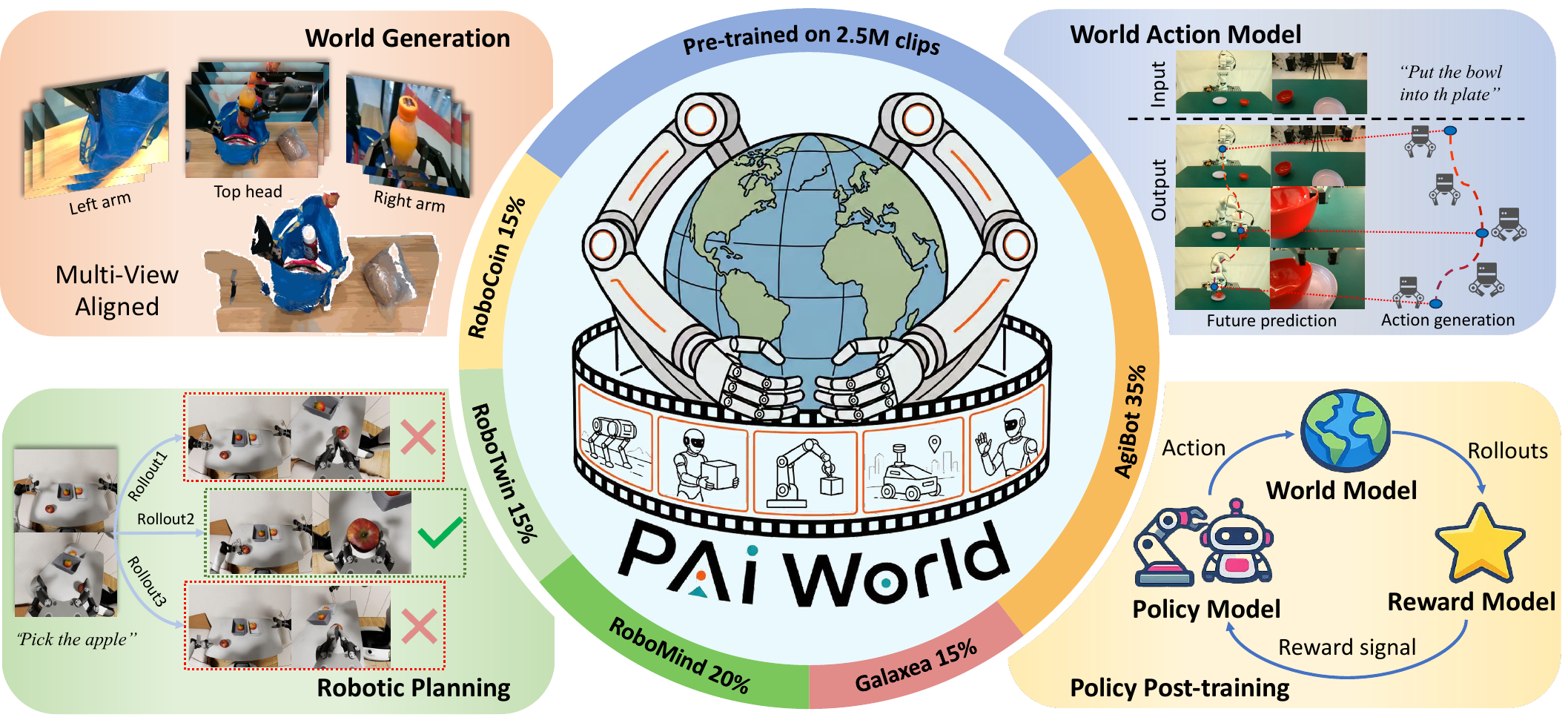}
	\caption{\methodname{} is a 3D-consistent multi-view world foundation model for robotic manipulation. Pre-trained on 2.5M multi-view video clips, \methodname{} serves as a versatile backbone for a range of downstream applications: multi-view world generation, world action modeling, robotic planning, and policy post-training. Across these settings, \methodname{} maintains cross-view 3D consistency, with coherent object placement, depth, and texture across all viewpoints, making its imagined rollouts physically plausible for embodied decision-making.}
	\label{fig:teaser}
\end{figure*}

World foundation models (WFMs) have rapidly progressed from compact latent-dynamics modules~\cite{ha2018world,hafner2020dreamer} to large-scale video generation systems capable of simulating complex physical environments~\cite{nvidia2025cosmos,nvidia2026cosmos3,liu2024sora,yang2024cogvideox,gao2024vista,kong2024hunyuanvideo,wan2025wan}. By learning to predict future visual observations conditioned on actions or text, WFMs serve as \emph{world simulators} that can be leveraged for model-based planning~\cite{wu2023daydreamer}, policy evolution~\cite{du2024learning}, and data synthesis in robotic learning~\cite{zhou2024robodreamer,huang2025ladiwm}. The Cosmos platform~\cite{nvidia2025cosmos}, in particular, has demonstrated that diffusion-transformer (DiT) architectures~\cite{peebles2023dit} trained on internet-scale video data can produce temporally coherent, physically plausible visual rollouts, making WFMs a promising backbone for embodied intelligence. Recent unified models such as Pelican-Unified~\cite{zhang2026pelican} further integrate world modeling with understanding, reasoning, and action within a single framework.

However, robotic manipulation systems are inherently \emph{multi-view}. Standard configurations employ wrist-mounted and egocentric cameras simultaneously to supply complementary geometric and semantic cues for policy learning~\cite{brohan2023rt1,brohan2023rt2,collaboration2023openx}. When a world model serves as a simulator for such a system, it must generate future observations across all viewpoints while maintaining strict 3D consistency: the same object must appear at geometrically compatible locations, with coherent depth and texture, across every view at every time step. Any breakdown in this consistency, whether cross-view object drift, depth contradictions, or texture misalignment, directly undermines the physical plausibility of imagined trajectories and propagates errors into downstream planning and control~\cite{wu2023daydreamer,chi2023diffusionpolicy}.

Existing approaches to multi-view world modeling fall short of this requirement. Single-view WFMs such as Cosmos~\cite{nvidia2025cosmos}, CogVideoX~\cite{yang2024cogvideox}, and Vista~\cite{gao2024vista} produce high-quality temporal predictions but are architecturally restricted to a single viewpoint. Methods that do handle multiple views, such as Genie~\cite{bruce2024genie} and iVideoGPT~\cite{wu2024ivideogpt}, typically concatenate tokens from different viewpoints along the sequence dimension without any explicit mechanism for cross-view geometric reasoning. This ``flat concatenation'' strategy treats multi-view tokens identically to temporal tokens, leaving the model to discover cross-view correspondences implicitly from data. Such implicit discovery grows increasingly unreliable as the number of viewpoints and the complexity of the scene scale up.

We trace the root cause of these failures to two fundamental deficiencies in existing multi-view approaches. \emph{First, they lack an explicit inter-view communication mechanism.} Flat concatenation provides no dedicated pathway for viewpoints to exchange information; each view's tokens attend to the entire sequence without distinguishing same-view from cross-view tokens, so the model must infer geometric correspondences implicitly. As a consequence, each viewpoint effectively generates in isolation, with no means to coordinate predictions or resolve cross-view conflicts. \emph{Second, they lack a 3D geometric prior.} Even with a communication pathway in place, the model receives no supervisory signal specifying what geometrically consistent 3D structure looks like. Absent such guidance, cross-view information exchange gravitates toward superficial shortcuts, such as matching color palettes or copying textures, rather than learning genuine 3D correspondences.

These two deficiencies call for two remedies that operate at distinct levels. At the \emph{architectural} level, the model needs an information pathway that lets viewpoints exchange features during generation; at the \emph{training-objective} level, it needs a geometric learning signal that steers what flows through this pathway toward 3D-consistent structure. Crucially, \emph{neither remedy alone is sufficient}, precisely because they act on different levels. An inter-view communication pathway without geometric supervision lets information flow but cannot guarantee that the exchanged content respects 3D geometry; in practice it learns shortcuts such as texture copying or uniform averaging. Conversely, a geometric prior without an inter-view pathway sharpens each view's 3D awareness in isolation, but the resulting constraints have no route to propagate across viewpoints, so cross-view inconsistencies persist. Only when both are present does the system achieve genuine multi-view 3D consistency: a pathway that carries information, and an objective that makes the information geometrically meaningful.

Based on this analysis, we present \textbf{\methodname{}}, a framework built on \emph{two technical pillars}, an architectural communication pathway and a geometric training objective, realized by three lightweight, modular components on the DiT backbone (\figref{fig:pipeline}). \emph{The first pillar, the inter-view pathway}, is realized by two cooperating components: \emph{Geometry-Aware Cross-View Attention} blocks, interleaved within the DiT, open the pathway for viewpoints to exchange features, while a shared \emph{Geometric Rotary Position Embedding} (Geo-RoPE) encodes camera ray directions and extrinsic poses into this attention via rotary position encoding~\cite{su2024rope,he2024cameractrl}, biasing the pathway to route information along geometrically corresponding tokens. \emph{The second pillar, the geometric objective}, is realized by the \emph{Latent 3D-REPA} that aligns intermediate DiT features with 3D-aware representations distilled from frozen 3D foundation models, which are trained under explicit 3D geometric supervision and thus encode genuine 3D structure, via the REPA framework~\cite{yu2024repa}, supplying the supervisory signal that makes the exchanged content 3D-consistent. In short, Cross-View Attention and Geo-RoPE let geometric information flow across views, while Latent 3D-REPA supervises that information to be 3D-consistent, so that only their combination produces coherent multi-view 3D structure. Our contributions are as follows:
\begin{itemize}
	\item We identify two fundamental deficiencies in existing multi-view world models, namely the lack of inter-view communication and the absence of a 3D geometric prior, and argue that addressing both jointly is necessary and sufficient for multi-view 3D consistency.
	\item We propose \methodname{}, which builds these two pillars from three plug-and-play components: Geometry-Aware Cross-View Attention and Geometric Rotary Position Embedding form the architectural pathway, while Latent 3D-REPA provides the geometric objective.
	\item We show that \methodname{} delivers state-of-the-art multi-view 3D consistency on robotic manipulation benchmarks: it ranks \textbf{1st} on the WorldArena benchmark (best overall EWMScore 72.31\%, with the best Motion Quality among all entries) and \textbf{2nd} on the AgiBot-Challenge2026 leaderboard (EWMScore 82.45\%), where it attains the \emph{best} Scene Consistency (90.41\%) among all entries. This improved consistency further translates directly into gains on downstream embodied tasks, including model-based robotic planning and world action model fine-tuning.
\end{itemize}

The remainder of this paper is organized as follows. \secref{sec:related-work} reviews related work on world models, multi-view generation, and 3D-aware representations. \secref{sec:methods} presents the \methodname{} framework in detail. \secref{sec:conclusion} summarizes our findings and discusses future directions.

\section{Related Work}
\label{sec:related-work}

\subsection{World Foundation Models for Physical AI}

The concept of learning internal models of the environment, commonly known as world models, has a long history in reinforcement learning and cognitive science~\cite{ha2018world}. Early approaches such as Dreamer~\cite{hafner2020dreamer} and DayDreamer~\cite{wu2023daydreamer} learn compact latent-state dynamics models that enable planning through imagined rollouts, demonstrating the viability of model-based learning for physical robot control. More recently, the success of large-scale generative models has motivated a paradigm shift toward \emph{world foundation models} (WFMs) that operate directly in pixel or video space. Sora~\cite{liu2024sora} demonstrated that video diffusion models can generate temporally coherent, physically plausible visual simulations. The Cosmos platform~\cite{nvidia2025cosmos} further formalized this direction by training DiT-based~\cite{peebles2023dit} video generation models on internet-scale data for physical AI applications, while Cosmos 3~\cite{nvidia2026cosmos3} extended it to an omnimodal framework unifying language, image, video, audio, and action modalities. CogVideoX~\cite{yang2024cogvideox}, HunyuanVideo~\cite{kong2024hunyuanvideo}, Stable Video Diffusion~\cite{blattmann2023svd}, and Wan~\cite{wan2025wan} achieved high-fidelity video generation across diverse domains. DIAMOND~\cite{alonso2024diamond} demonstrated that diffusion-based world models can capture fine visual details critical for decision-making. Pelican-Unified~\cite{zhang2026pelican} advocates a unified paradigm integrating world modeling with understanding, reasoning, and action for embodied intelligence. However, these models are fundamentally single-view: they generate one coherent video stream without explicit mechanisms for multi-view consistency.

Interactive world models represent another important direction. UniSim~\cite{yang2023unisim} learns real-world simulators from diverse data sources, Genie~\cite{bruce2024genie} and Genie 2~\cite{parkerholder2024genie2} enable interactive environment generation from single images, and iVideoGPT~\cite{wu2024ivideogpt} scales autoregressive world models to complex environments. In the robotic manipulation domain, GR-2~\cite{cheang2024gr2} builds a generative video-language-action model with web-scale knowledge, IRASim~\cite{zhu2025irasim} learns action-conditioned video simulators from real robot data, EnerVerse~\cite{huang2025enerverse} envisions embodied future spaces for manipulation planning, and LaDi-WM~\cite{huang2025ladiwm} employs latent diffusion-based world models for predictive manipulation. AgiBot World~\cite{bu2025agibotworld} provides a large-scale multi-view manipulation platform, and WorldArena~\cite{shang2026worldarena} offers a unified benchmark for evaluating embodied world models. While some of these systems support multiple viewpoints via token concatenation, none introduces explicit cross-view geometric reasoning, a critical limitation for robotic applications, where 3D consistency directly governs policy quality.

\subsection{Multi-View and 3D-Aware Visual Generation}

Multi-view generation has been extensively studied in the context of 3D content creation from single images. Zero-1-to-3~\cite{liu2023zero123} pioneered viewpoint-conditioned diffusion by fine-tuning a 2D diffusion model with relative camera transformations. SyncDreamer~\cite{liu2023syncdreamer} introduced synchronized multi-view denoising with 3D-aware attention to ensure geometric consistency across generated views. MVDream~\cite{shi2023mvdream} extended multi-view diffusion to text-conditioned 3D generation, while MVDiffusion~\cite{tang2023mvdiffusion} enabled holistic multi-view image generation with correspondence-aware attention. SV3D~\cite{voleti2024sv3d} leveraged latent video diffusion for orbital multi-view synthesis, and SV4D~\cite{xie2024sv4d} further extended this to dynamic 3D content with multi-frame and multi-view consistency. CAT3D~\cite{gao2024cat3d} demonstrated that multi-view diffusion models can create high-quality 3D assets from arbitrary inputs.

Our work differs from this line in three respects, which together define a regime these methods do not target. \emph{First, scene scope}: the above methods are object-centric, generating views of a single foreground object against a clean background, whereas robotic manipulation requires modeling cluttered scenes with a manipulator, objects, and a dynamic background. \emph{Second, dynamics}: they synthesize static objects or short orbital trajectories, while a world model must roll out temporally evolving dynamics driven by text or actions. \emph{Third, camera configuration}: they assume dense, smoothly-varying viewpoints around an object, whereas robotic rigs provide a few fixed, wide-baseline cameras (egocentric, eye-to-hand, wrist) with little view overlap, where implicit correspondence learning is far harder. \methodname{} is designed for this dynamic, wide-baseline, scene-level regime, and injects geometry through an explicit communication pathway and a 3D supervisory objective rather than relying on dense view sampling.

Camera control in video generation has been explored by CameraCtrl~\cite{he2024cameractrl}, which represents camera poses using Pl\"ucker ray coordinates and injects them into video diffusion models. ViewCrafter~\cite{yu2024viewcrafter} further tames video diffusion models for high-fidelity novel view synthesis. These works provide technical foundations for camera-aware generation, but focus on single-view camera trajectory control rather than multi-view consistency.

\subsection{3D Representations and Geometric Reconstruction}

Recent advances in geometric 3D vision have produced powerful tools for evaluating and enforcing 3D consistency. Neural Radiance Fields (NeRF)~\cite{mildenhall2020nerf} and 3D Gaussian Splatting~\cite{kerbl2023gaussiansplatting} established the foundation for differentiable 3D scene representations. DUSt3R~\cite{wang2024dust3r} and its successor MASt3R~\cite{leroy2024mast3r} learn to predict dense 3D point maps from image pairs without requiring known camera parameters, enabling direct measurement of cross-view geometric consistency. The Depth Anything series~\cite{yang2024depthanything,yang2024depthanythingv2,lin2025depthanything3} provides foundation models for monocular depth estimation that capture robust 3D-aware features from large-scale training. VGGT~\cite{wang2025vggt} further advances this direction by grounding visual geometry in a unified transformer that jointly predicts camera poses, depth, and 3D point maps from arbitrary image collections. These models encode rich geometric priors that can serve as supervision targets for world models seeking 3D consistency.

The REPA framework~\cite{yu2024repa} introduced the idea of aligning intermediate representations of diffusion transformers with those of pretrained encoders, demonstrating accelerated training and improved generation quality for image diffusion. Our Latent 3D-REPA extends this principle from 2D image generation to multi-view video world models, using 3D-aware features from Depth Anything 3 as the alignment target to inject geometric consistency directly into the diffusion process.

\section{Method}
\label{sec:methods}

\begin{figure*}[t]
	\centering
	\includegraphics[width=\textwidth]{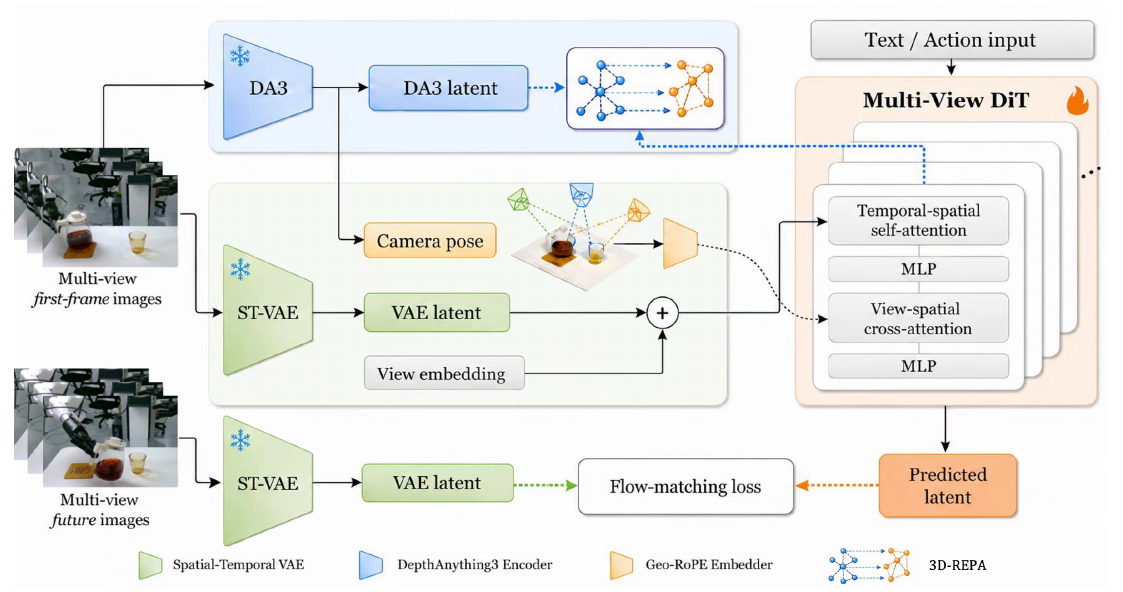}
	\caption{Overview of the \methodname{} framework. Built on a DiT-based flow matching backbone, \methodname{} rests on two technical pillars realized by three components. \emph{Pillar 1, the inter-view pathway} (two components): (1)~Geometry-Aware Cross-View Attention blocks open an explicit communication pathway across views, and (2)~Geometric Rotary Position Embedding (Geo-RoPE) encodes camera ray directions and extrinsic poses into this attention, biasing it toward geometrically corresponding tokens. \emph{Pillar 2, the geometric objective} (one component): (3)~Latent 3D-REPA aligns intermediate DiT representations with 3D-aware features from a frozen 3D foundation model (Depth Anything 3), supplying the supervisory signal. The first two components build the pathway, while Latent 3D-REPA ensures the information flowing through it is 3D-consistent.}
	\label{fig:pipeline}
\end{figure*}

We present \methodname{}, a framework for injecting 3D consistency into flow-matching world foundation models for multi-view robotic manipulation. As argued in \secref{sec:introduction}, multi-view 3D consistency rests on two technical pillars: an architectural pathway for inter-view communication and a training objective that enforces 3D-consistent content. \methodname{} builds these two pillars from three components on a DiT-based flow matching backbone (\figref{fig:pipeline}). The pathway is formed by two cooperating components: Geometry-Aware Cross-View Attention (\secref{sec:cross-view-attn}), which opens the inter-view pathway, and Geometric Rotary Position Embedding (\secref{sec:camera-pe}), which shapes this pathway with camera geometry. The objective is supplied by a single component, Latent 3D-REPA (\secref{sec:3d-repa}). We first formalize the problem setting (\secref{sec:problem}), then describe each component and analyze why both pillars must be present simultaneously (\secref{sec:synergy}).

\subsection{Problem Formulation}
\label{sec:problem}

Consider a robotic manipulation system equipped with $V$ cameras, each providing a video stream. At time step $t$, the system observes a set of images $\{I_t^v\}_{v=1}^{V}$ with associated camera intrinsics $\{\mathbf{K}^v\}_{v=1}^{V}$ and extrinsics $\{[\mathbf{R}^v \mid \mathbf{t}^v]\}_{v=1}^{V} \in \mathrm{SE}(3)$. Given a conditioning signal $c$ (text description or action sequence) and context frames $\{I_{1:t_0}^{1:V}\}$, the goal of multi-view video generation is to model the conditional distribution:
\begin{equation}
\label{eq:mv-gen}
p_\theta\!\left(\{I_{t_0+1:T}^{v}\}_{v=1}^{V} \;\middle|\; \{I_{1:t_0}^{v}\}_{v=1}^{V},\, \{\mathbf{K}^v, \mathbf{R}^v, \mathbf{t}^v\}_{v=1}^{V},\, c\right),
\end{equation}
where $T$ is the prediction horizon. Beyond per-view fidelity, the generated multi-view video must satisfy a \emph{multi-view 3D consistency} requirement: at every time step, the views should admit a coherent 3D explanation. Formally, there exists a consistent 3D scene $\mathcal{S}_t$ such that all views $\{I_t^v\}_{v=1}^V$ can be obtained by rendering $\mathcal{S}_t$ from their respective camera poses; equivalently, points corresponding to the same 3D location across views must respect the underlying epipolar geometry. This requirement is what existing single-view and flat-concatenation models fail to enforce, and it motivates the two pillars of our design.

\subsection{Flow Matching DiT for Video Generation}

We adopt a Diffusion Transformer (DiT)~\cite{peebles2023dit} trained with a flow matching objective~\cite{lipman2023flow,esser2024sd3} operating in the latent space of a pretrained video VAE. Following Wan2.1~\cite{wan2025wan}, we use its spatial-temporal VAE to compress each video both spatially and temporally into a compact latent representation, which substantially reduces the token count and makes multi-view video modeling tractable. Given an input video, the VAE encoder produces a latent representation $\feat_0 \in \R^{T \times H \times W \times C}$. The model learns a velocity field $u_\theta(\feat_s, s)$ that transports samples from a noise distribution toward the data distribution along a linear interpolation path:
\begin{equation}
\label{eq:flow-matching}
\feat_s = (1 - s) \feat_0 + s \boldsymbol{\epsilon}, \quad \boldsymbol{\epsilon} \sim \mathcal{N}(\mathbf{0}, \mathbf{I}),
\end{equation}
where $s \in [0, 1]$ is the flow timestep. The training objective minimizes:
\begin{equation}
\label{eq:flow-loss}
\lossdiff = \E_{s, \boldsymbol{\epsilon}}\left[\|u_\theta(\feat_s, s) - (\boldsymbol{\epsilon} - \feat_0)\|_2^2\right].
\end{equation}

\paragraph{Conditioning Signal Injection.}
The conditioning signal $c$ is injected into each DiT block via Adaptive Layer Normalization (AdaLN)~\cite{peebles2023dit}. For text-conditioned generation, $c$ is a text embedding that modulates the scale and shift parameters of layer normalization, globally steering the generation toward the described scene. For action-conditioned generation, rather than representing robot actions as raw vectors, we follow EVAC~\cite{huang2025evac} and render actions into spatial \emph{action maps} that are concatenated with the noisy latent along the channel dimension. This spatial representation preserves the geometric structure of the action (e.g., end-effector trajectory projected into each camera view), enabling the model to ground action semantics in pixel space rather than learning an implicit mapping from abstract action vectors.

\paragraph{Multi-View Token Concatenation.}
For multi-view generation, a naive approach concatenates all view tokens along the sequence dimension, yielding $\feat_0^{\text{concat}} \in \R^{(V \cdot T) \times H \times W \times C}$. While the standard temporal self-attention operates over this concatenated sequence, it treats multi-view tokens identically to temporal tokens without any geometric inductive bias. The model must discover cross-view correspondences entirely from data, which is insufficient for consistent 3D generation.

\subsection{Geometric Rotary Position Embedding}
\label{sec:camera-pe}

To encode 3D geometric information into the attention mechanism, we introduce a \emph{Geometric Rotary Position Embedding} (Geo-RoPE) that separately encodes pixel-level ray directions and view-level camera poses via rotary position encoding~\cite{su2024rope}.

\paragraph{Dual-Component Design.}
For each attention head with dimension $d$, we split the query and key vectors into two equal subspaces: a \emph{ray subspace} of dimension $d_r = d/2$ and a \emph{pose subspace} of dimension $d_p = d/2$. Each subspace receives its own geometric encoding via RoPE.

\paragraph{Ray Component.}
For each token at spatial location $(h, w)$ in view $v$, we compute the world-space ray direction by unprojecting the pixel through the camera intrinsics $\mathbf{K}^v$ and rotating by the inverse of the camera rotation $(\mathbf{R}^v)^{-1}$:
\begin{equation}
\label{eq:ray-dir}
\mathbf{d}^v(h,w) = \text{normalize}\left((\mathbf{R}^v)^\top \cdot (\mathbf{K}^v)^{-1} \begin{pmatrix} h + 0.5 \\ w + 0.5 \\ 1 \end{pmatrix}\right) \in \R^3.
\end{equation}
The 3D ray direction is cyclically expanded to fill the $d_r/2$ frequency slots and used as position coordinates for RoPE rotation on the ray subspace of $\mathbf{q}$ and $\mathbf{k}$.

\paragraph{Pose Component.}
For each view $v$, we extract a 12-dimensional pose feature vector that captures the full camera geometry:
\begin{equation}
\label{eq:pose-feat}
\mathbf{e}^v = [\underbrace{\text{yaw}, \text{pitch}, \text{roll}}_{\text{Euler angles}},\; \underbrace{\mathbf{t}^v}_{\text{translation}},\; \underbrace{-(\mathbf{R}^v)^\top \mathbf{t}^v}_{\text{camera position}},\; \underbrace{(\mathbf{R}^v)^\top \mathbf{e}_z}_{\text{optical axis}}] \in \R^{12},
\end{equation}
where $\mathbf{e}_z = [0,0,1]^\top$. This pose vector is shared across all spatial positions within a view and applied via RoPE to the pose subspace.

\paragraph{Split-RoPE Application.}
The complete Geo-RoPE operates as:
\begin{equation}
\label{eq:geo-rope}
\begin{aligned}
\mathbf{q}_{\text{ray}},\, \mathbf{q}_{\text{pose}} &= \text{split}(\mathbf{q},\; [d_r, d_p]) \\
\tilde{\mathbf{q}}_{\text{ray}} &= \text{RoPE}(\mathbf{q}_{\text{ray}},\; \mathbf{d}^v(h,w)) \\
\tilde{\mathbf{q}}_{\text{pose}} &= \text{RoPE}(\mathbf{q}_{\text{pose}},\; \mathbf{e}^v) \\
\tilde{\mathbf{q}} &= [\tilde{\mathbf{q}}_{\text{ray}};\; \tilde{\mathbf{q}}_{\text{pose}}]
\end{aligned}
\end{equation}
The same operation applies to keys. This design ensures that the ray subspace captures fine-grained pixel-level geometric correspondences (tokens viewing the same 3D point receive similar rotations), while the pose subspace encodes view-level identity (tokens from the same camera share identical pose rotations). Separating the two prevents interference between the spatially-varying ray signal and the spatially-uniform pose signal.

\subsection{Geometry-Aware Cross-View Attention}
\label{sec:cross-view-attn}

Standard temporal self-attention in each DiT block operates independently per view (after merging $V$ into the batch dimension), providing no explicit cross-view interaction. To open the inter-view communication pathway, we introduce two complementary attention mechanisms: dedicated Cross-View Attention blocks and periodic spatial-concat self-attention.

\paragraph{Multi-View Self-Attention Blocks.}
At selected DiT layers, we insert a dedicated \emph{Cross-View Attention} sub-block. For each temporal frame $t$ (distinct from the flow timestep $s$), let $\{\mathbf{Z}_t^v\}_{v=1}^{V}$ denote the feature maps from all views, where $\mathbf{Z}_t^v \in \R^{(H \cdot W) \times D}$. Each view's queries and keys are first projected and then rotated by Geo-RoPE \emph{using that view's own camera geometry}, so that the rotation applied to view $v$ depends on the rays and pose of camera $v$:
\begin{equation}
\label{eq:cross-view-attn}
\begin{aligned}
\tilde{\mathbf{Q}}_t^v &= \mathrm{GeoRoPE}_v\!\left(\mathbf{W}_Q \mathbf{Z}_t^v\right), \qquad
\tilde{\mathbf{K}}_t^v = \mathrm{GeoRoPE}_v\!\left(\mathbf{W}_K \mathbf{Z}_t^v\right), \qquad
\mathbf{V}_t^v = \mathbf{W}_V \mathbf{Z}_t^v, \\
\hat{\mathbf{Z}}_t^v &= \mathbf{Z}_t^v + \text{gate} \cdot \mathrm{softmax}\!\left(\frac{\tilde{\mathbf{Q}}_t^v \, [\tilde{\mathbf{K}}_t^1; \ldots; \tilde{\mathbf{K}}_t^V]^\top}{\sqrt{d}}\right) [\mathbf{V}_t^1; \ldots; \mathbf{V}_t^V],
\end{aligned}
\end{equation}
where $\mathrm{GeoRoPE}_v$ denotes the split ray-pose rotary encoding of \eqnref{eq:geo-rope} instantiated with the intrinsics $\mathbf{K}^v$ and extrinsics $[\mathbf{R}^v \mid \mathbf{t}^v]$ of view $v$, and $[\,\cdot\,;\,\cdot\,]$ concatenates the per-view keys (values) along the token dimension. Because each view is rotated by its own geometry, the query of view $v$ and the key of view $v'$ attain a high inner product precisely when their tokens observe the same 3D point, so geometrically corresponding tokens across views naturally receive higher attention weights. The gate is initialized to zero via AdaLN-Zero, preserving the pretrained single-view model at initialization.

\paragraph{Spatial-Concat Self-Attention.}
Periodically, in place of per-view temporal attention, we flatten the view and spatial dimensions into a single token axis of length $V \cdot H \cdot W$ and perform joint spatio-view self-attention over all these tokens simultaneously. This provides a broader receptive field in which each token can attend to all spatial positions across all views within the same temporal context, complementing the dedicated cross-view blocks.

\paragraph{Why the Pathway Alone Is Insufficient.}
Cross-View Attention and Geo-RoPE together establish the architectural pathway along which viewpoints exchange information, biased toward geometrically corresponding tokens. Yet an architectural bias only shapes \emph{how} information flows; it does not dictate \emph{what content} is 3D-consistent. Without an explicit geometric objective, the pathway can still settle into trivial shortcuts, such as copying textures across views or averaging features, that minimize the generation loss while violating true 3D structure. This motivates the geometric learning signal introduced next.

\subsection{Latent 3D-REPA (3D Geometric Prior)}
\label{sec:3d-repa}

To provide the geometric learning signal for cross-view interaction, we introduce Latent 3D-REPA, a token-relation distillation objective that aligns the DiT's intermediate representations with features from frozen 3D foundation models.

\paragraph{3D-Aware Feature Extraction.}
We employ Depth Anything 3~\cite{lin2025depthanything3} as our 3D-aware feature extractor. Crucially, Depth Anything 3 is trained under direct 3D geometric supervision, predicting depth, point maps, and camera poses across multiple views; its intermediate features therefore internalize explicit 3D structure rather than merely 2D appearance. This property is what makes aligning to its features a meaningful source of geometric supervision: distilling its representations transfers 3D knowledge that the generative backbone, trained only on a reconstruction objective, does not otherwise acquire. For each multi-view frame set $\{I_t^v\}_{v=1}^V$, Depth Anything 3 produces dense features that encode rich geometric priors including depth, 3D point maps, and camera-relative spatial structure. In addition, it recovers camera extrinsics and intrinsics that are used by Geo-RoPE and for 3D point map reconstruction.

\paragraph{Token Relation Distillation.}
Rather than directly regressing 3D features token-by-token, we distill the \emph{relational structure} between tokens. At a selected intermediate layer $\ell$ of the DiT, we extract the feature map $\mathbf{H}_\ell$ and project it to the VGGT feature dimension via a lightweight 3D convolutional projector $g_\phi$, yielding per-token features $\mathbf{F}^{\text{DiT}} = g_\phi(\mathbf{H}_\ell)$; the corresponding frozen features from Depth Anything 3 are denoted $\mathbf{F}^{\text{DA3}}$. We supervise the \emph{pairwise relations} among these tokens rather than their absolute values, since relational structure is invariant to the feature-space discrepancy between the two encoders.

Computing the full token-token similarity matrix is prohibitively expensive ($N{=}V\!\cdot\!H\!\cdot\!W$ tokens per frame, over $T$ frames). We therefore estimate the relations through \emph{anchor sampling}: a random subset of $K$ tokens is drawn as anchors, and we measure the cosine similarity between every token and each anchor. For a token set $\mathbf{F} = \{\mathbf{f}_i\}_{i=1}^{M}$ with sampled anchor indices $\mathcal{A} \subset \{1,\dots,M\}$, $|\mathcal{A}| = K$, we define the sampled similarity matrix $\mathbf{S}(\mathbf{F}) \in \R^{M \times K}$ as
\begin{equation}
\label{eq:sampled-sim}
\mathbf{S}(\mathbf{F})_{i,a} = \frac{\mathbf{f}_i^\top \mathbf{f}_{a}}{\|\mathbf{f}_i\| \, \|\mathbf{f}_{a}\|}, \quad a \in \mathcal{A}.
\end{equation}
The distillation loss aligns this relational structure at two granularities, applying the operator $\mathbf{S}(\cdot)$ of \eqnref{eq:sampled-sim} to two different token sets:
\begin{equation}
\label{eq:3d-repa}
\lossrepa = \mathcal{L}_{\text{spatial}} + \mathcal{L}_{\text{temporal}}.
\end{equation}
The \emph{spatial} term operates within each frame: let $\mathbf{F}_{\text{frame}}$ denote the $N$ tokens of a single frame (across all views and spatial positions), from which $K_s$ anchors are sampled. Writing $\mathbf{S}_{\text{intra}} := \mathbf{S}(\mathbf{F}_{\text{frame}}) \in \R^{N \times K_s}$, the term aligns the DiT and Depth Anything 3 relations:
\begin{equation}
\mathcal{L}_{\text{spatial}} = \text{SmoothL1}\!\left(\mathbf{S}_{\text{intra}}^{\text{DiT}},\; \mathbf{S}_{\text{intra}}^{\text{DA3}}\right).
\end{equation}
The \emph{temporal} term operates over the entire clip: let $\mathbf{F}_{\text{clip}}$ denote the full set of $T\!\cdot\!N$ tokens, from which $K_t$ anchors are sampled across all frames. Writing $\mathbf{S}_{\text{inter}} := \mathbf{S}(\mathbf{F}_{\text{clip}}) \in \R^{(T \cdot N) \times K_t}$,
\begin{equation}
\mathcal{L}_{\text{temporal}} = \text{SmoothL1}\!\left(\mathbf{S}_{\text{inter}}^{\text{DiT}},\; \mathbf{S}_{\text{inter}}^{\text{DA3}}\right).
\end{equation}
Both $\mathbf{S}_{\text{intra}}$ and $\mathbf{S}_{\text{inter}}$ are instances of the same sampled-similarity operator $\mathbf{S}(\cdot)$, differing only in their token set and anchor count: $\mathbf{S}_{\text{intra}}$ captures intra-frame geometric relations (within a single time step, across views and space), while $\mathbf{S}_{\text{inter}}$ captures cross-frame relations that span the temporal dimension. Stochastic anchor sampling reduces the cost from quadratic to $O(MK)$ while retaining an effective gradient signal.

\paragraph{Why Geometric Priors Alone Are Insufficient.}
Latent 3D-REPA encourages the DiT's token relations to mirror those of a geometry-aware encoder, providing an explicit signal of correct 3D structure. Yet this per-frame prior cannot enforce \emph{cross-view} consistency on its own: each view improves its 3D awareness in isolation, but without an inter-view pathway the views cannot coordinate to produce geometrically compatible outputs.

\subsection{Joint Mechanism}
\label{sec:synergy}

Neither pillar suffices in isolation, as motivated above; their value emerges from coupling. When both are present, they form a reinforcing loop:
\begin{enumerate}
    \item \emph{The pathway (architecture).} Geometry-Aware Cross-View Attention opens the pathway through which views exchange information, and Geo-RoPE biases this pathway to route information along geometrically corresponding tokens, while also placing all views in a shared 3D coordinate frame.
    \item \emph{The objective (supervision).} Latent 3D-REPA ensures the exchanged content is geometrically meaningful, aligning the DiT's token relations with a 3D-aware prior that spans all views; because Geo-RoPE has fixed a common reference frame, these supervised constraints propagate coherently across views rather than collapsing into per-view shortcuts.
\end{enumerate}
The pathway carries information and the objective makes it 3D-consistent; because each addresses a deficiency the other cannot, their combination yields a non-additive improvement in 3D consistency.

\subsection{Training Objective}
\label{sec:training}

The total training objective combines the flow matching loss with the Latent 3D-REPA distillation loss:
\begin{equation}
\label{eq:total-loss}
\losstotal = \lossdiff + \lambda \cdot \lossrepa,
\end{equation}
where $\lossdiff$ is the flow matching velocity prediction loss from \eqnref{eq:flow-loss} and $\lambda = 0.5$ balances generation quality with 3D alignment.

The Depth Anything 3 encoder is kept frozen throughout training, serving as a fixed 3D prior. Cross-View Attention blocks are initialized with AdaLN-Zero gating (gate$=0$ at initialization) so that the pretrained backbone weights are exactly preserved at step zero, and the new modules gradually contribute as training progresses.

\section{Experiments}
\label{sec:experiments}

We evaluate \methodname{} on two generation paradigms: action-conditioned video generation and text-conditioned multi-view generation. We first describe our implementation details, then present quantitative comparisons against state-of-the-art baselines across three benchmarks.

\subsection{Implementation Details}
\label{sec:impl}

\paragraph{Base Model.}
We build \methodname{} upon Cosmos-Predict2.5~\cite{nvidia2025cosmospredict25}, a flow-matching diffusion transformer (DiT) world foundation model operating in the latent space of a pretrained video VAE. Our full model has approximately 14B parameters. We adopt Cosmos-Reason1~\cite{nvidia2025cosmosreason1}, a Physical AI vision-language model, as the text embedder to provide physically-grounded conditioning. The Geometry-Aware Cross-View Attention blocks and Geo-RoPE modules are inserted into the pretrained backbone, while the REPA projection heads are randomly initialized.

\paragraph{Dataset.}
We curate a large-scale multi-view robotic manipulation dataset of approximately 2.5M video clips from five sources: AgiBot-World~\cite{bu2025agibotworld} (35\%), RoboMIND~\cite{wu2024robomind} (20\%), Galaxea~\cite{jiang2025galaxea} (15\%), RoboTwin~\cite{mu2025robotwin} (15\%), and RoboCOIN~\cite{wu2025robocoin} (15\%). These datasets provide multi-camera video streams of robotic manipulation, accompanied by text descriptions or action annotations. Together they span diverse embodiments, manipulation tasks, and camera configurations, offering broad coverage for training a generalizable multi-view world model. For action-conditioned generation, we further fine-tune on task-specific data from the AgiBot-Challenge2026 and WorldArena benchmarks.

\paragraph{Training Configuration.}
Training proceeds for a total of 30,000 iterations with a batch size proportional to the GPU count. We use the AdamW optimizer with a cosine learning rate schedule. The learning rate warms up linearly over the first 3,000 iterations to peak value $3 \times 10^{-5}$, then decays following a cosine schedule. 
All training experiments are conducted with NVIDIA H200 GPU and takes approximately 30k GPU-hours.
The REPA alignment loss weight is set to $\lambda = 0.5$. The 3D foundation model encoder (Depth Anything 3~\cite{lin2025depthanything3}) is kept frozen throughout training.

\subsection{Action-Conditioned Video Generation}
\label{sec:exp-action}

We first evaluate \methodname{} in the action-conditioned setting, where the model generates future observations given a sequence of robot actions. This setting directly measures the model's utility as a world simulator for robotic planning and control. We report results on two benchmarks: WorldArena and AgiBot-Challenge2026.

\subsubsection{WorldArena Benchmark}

The WorldArena benchmark~\cite{shang2026worldarena} provides a comprehensive evaluation suite with seven fine-grained metrics that decompose world model quality into distinct dimensions. Results are presented in \tabref{tab:action-worldarena}.

\begin{table*}[!t]
	\centering
	\caption{Action-conditioned generation results on the WorldArena benchmark. Best results are in \textbf{bold}, second best are \underline{underlined}.}
	\label{tab:action-worldarena}
	\renewcommand{\arraystretch}{1.15}
	\resizebox{\textwidth}{!}{%
		\begin{tabular}{lccccccc}
			\toprule
			Method & EWMScore $\uparrow$ & \shortstack{Visual\\Quality} $\uparrow$ & \shortstack{Motion\\Quality} $\uparrow$ & \shortstack{Content\\Consistency} $\uparrow$ & \shortstack{Physics\\Adherence} $\uparrow$ & \shortstack{3D\\Accuracy} $\uparrow$ & Controllability $\uparrow$ \\
			\midrule	
            WorldScape v0.2 & 68.32 & 62.65 & 42.34 & \underline{65.18} & \textbf{73.29} & 96.28 & 87.59 \\
			SparkWorld & 68.65 & \textbf{65.04} & 59.71 & 59.59 & 58.59 & 92.54 & 87.26 \\            MotuBrain~\cite{motubrainteam2026motubrainadvancedworldaction}& 69.06 & \underline{63.98} & 64.58 & 59.51 & 58.50 & 91.48 & 85.84 \\
            Z-WM & 69.59 & 61.63 & 49.97 & \textbf{65.51} & \underline{69.81} & \underline{97.16} & \underline{89.28} \\
            GenieEnvisioner-Sim2.0-2B~\cite{liao2025genieenvisioner} & 69.59 & 60.99 & 62.16 & 60.12 & 66.11 & 95.06 & 85.88 \\
            Pelican-Unify~\cite{zhang2026pelican} & 70.38 & 63.60 & 61.73 & 60.41 & 63.98 & \textbf{97.65} & 87.60 \\
            FlowWAM-FiveAges& 72.00 & 63.43 & 79.45 & 57.43 & 59.80 & 91.60 & 88.16 \\
            BWM-Fast & 72.15 & 62.79 & 78.79 & 58.30 & 61.18 & 91.53 & 88.58 \\
            UNIS & \underline{72.16} & 60.85 & \textbf{81.60} & 56.44 & 61.56 & 91.16 & \textbf{90.19} \\
			\textbf{\methodname{} (Ours)} & \textbf{72.31} & 63.04 & \underline{80.45} & 57.85 & 61.66 & 91.51 & 87.16 \\
			\bottomrule
		\end{tabular}%
	}
\end{table*}

\paragraph{Evaluation Metrics.}
\begin{itemize}[nosep]
	\item \textbf{EWMScore}: Overall world model quality score.
	\item \textbf{Visual Quality}: Perceptual quality of individual generated frames.
	\item \textbf{Motion Quality}: Smoothness and realism of temporal dynamics.
	\item \textbf{Content Consistency}: Semantic and appearance coherence across frames and views.
	\item \textbf{Physics Adherence}: Physical plausibility of object interactions and dynamics.
	\item \textbf{3D Accuracy}: Geometric correctness of cross-view 3D structure.
	\item \textbf{Controllability}: Fidelity of generated video to the input action commands.
\end{itemize}

\begin{figure*}[!t]
	\centering
	\includegraphics[width=\textwidth]{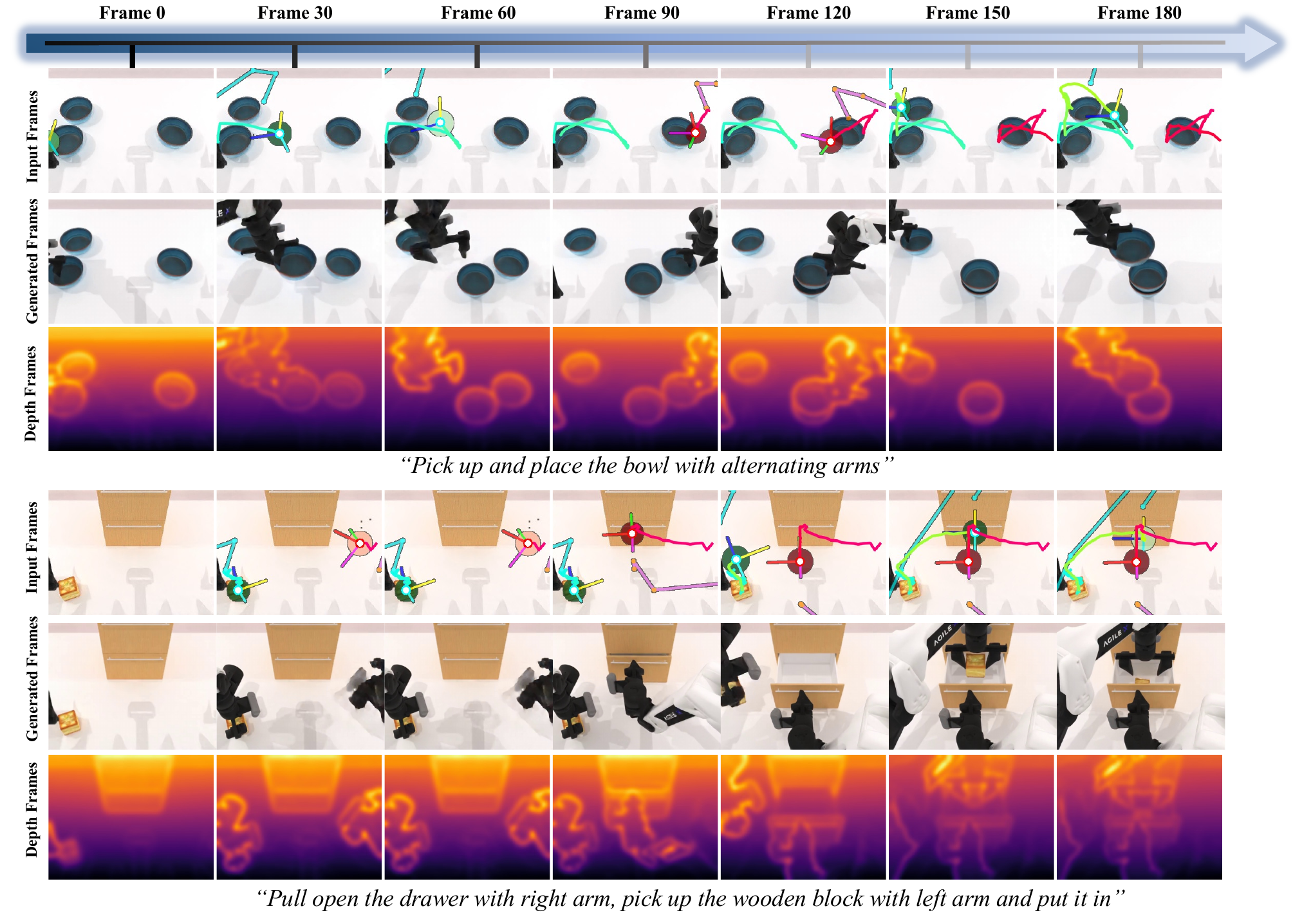}
	\caption{Qualitative results on the WorldArena benchmark. Each row shows a future rollout generated from an initial observation and a commanded action sequence. \methodname{} produces physically plausible dynamics, with object interactions and motion that respect the action commands, while keeping the scene layout stable over long horizons.}
	\label{fig:worldarena-visual}
\end{figure*}

\begin{figure*}[!t]
	\centering
	\includegraphics[width=\textwidth]{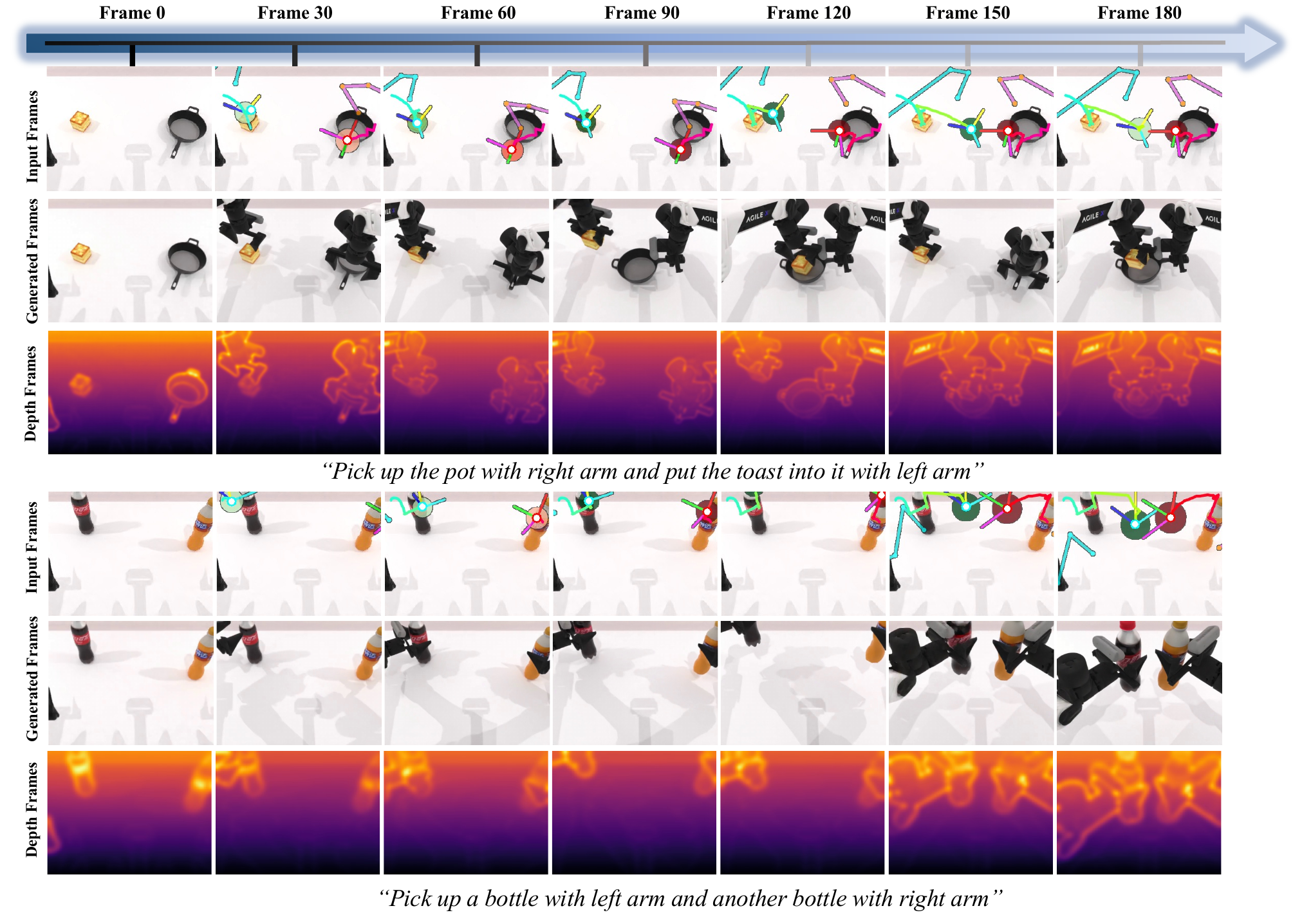}
	\caption{Additional qualitative results on the WorldArena benchmark. \methodname{} generates temporally coherent rollouts across diverse scenes and action sequences, maintaining consistent object appearance and physically plausible motion throughout the predicted horizon.}
	\label{fig:worldarena-visual2}
\end{figure*}

As reported in \tabref{tab:action-worldarena}, \methodname{} ranks \textbf{1st} on the WorldArena benchmark, achieving the best overall EWMScore of 72.31 and edging out the strongest competing entries (UNIS, 72.16; BWM-Fast, 72.15). The EWMScore aggregates seven fine-grained dimensions, and our top ranking reflects a consistently strong profile across them rather than a single outlier: in particular, \methodname{} attains the second-best \emph{Motion Quality} (80.45, just behind UNIS at 81.60), reflecting the temporally coherent, physically plausible dynamics that action-conditioned rollouts demand. Crucially, the competing methods each excel on only a narrow subset of dimensions: WorldScape v0.2 tops Physics Adherence but collapses on Motion Quality (42.34), while UNIS leads Motion Quality and Controllability yet trails on Content Consistency and 3D Accuracy. \methodname{} is the only entry that remains near the top across every axis simultaneously, and this all-around balance, rather than dominance on any single metric, is precisely what the aggregate EWMScore rewards and what a deployable world simulator demands: it must render faithfully, move plausibly, respect physics, preserve 3D structure, and obey action commands all at once. \figref{fig:worldarena-visual} presents representative rollouts, where the generated observations track the commanded actions and preserve physically plausible scene dynamics over time.

\subsubsection{AgiBot-Challenge2026 Benchmark}

The AgiBot-Challenge2026 benchmark evaluates action-conditioned world models on robotic manipulation tasks with four metrics. Results are reported in \tabref{tab:action-agibot}.

\begin{table}[!t]
	\centering
	\small
	\caption{Action-conditioned generation results on the AgiBot-Challenge2026 benchmark. Best results are in \textbf{bold}, second best are \underline{underlined}.}
	\label{tab:action-agibot}
	\renewcommand{\arraystretch}{1.15}
	\begin{tabular}{lcccc}
		\toprule
		Team & EWMScore $\uparrow$ & PSNR $\uparrow$ & Scen Cons. $\uparrow$ & nDTW $\uparrow$ \\
		\midrule
		NeoVerse-ABot & \textbf{0.829} & \textbf{0.6246} & 0.8974 & \textbf{0.9651} \\
		Loop & 0.8241 & \underline{0.6207} & \underline{0.9024} & 0.9492 \\
		Wild Path & 0.8232 & -- & -- & -- \\
		VIPL-GENUN & 0.8195 & -- & -- & -- \\
		\textbf{\methodname{} (Ours)} & \underline{0.8245} & 0.6161 & \textbf{0.9041} & \underline{0.9531} \\
		\bottomrule
	\end{tabular}
\end{table}

\paragraph{Evaluation Metrics.}
\begin{itemize}[nosep]
	\item \textbf{EWMScore}: Overall world model quality score.
	\item \textbf{PSNR}: Peak Signal-to-Noise Ratio measuring reconstruction fidelity.
	\item \textbf{Scene Consistency}: Temporal semantic coherence measured by DINOv2 feature similarity.
	\item \textbf{nDTW}: Normalized Dynamic Time Warping measuring trajectory alignment between generated and ground-truth sequences.
\end{itemize}

\begin{figure*}[!t]
	\centering
	\includegraphics[width=\textwidth]{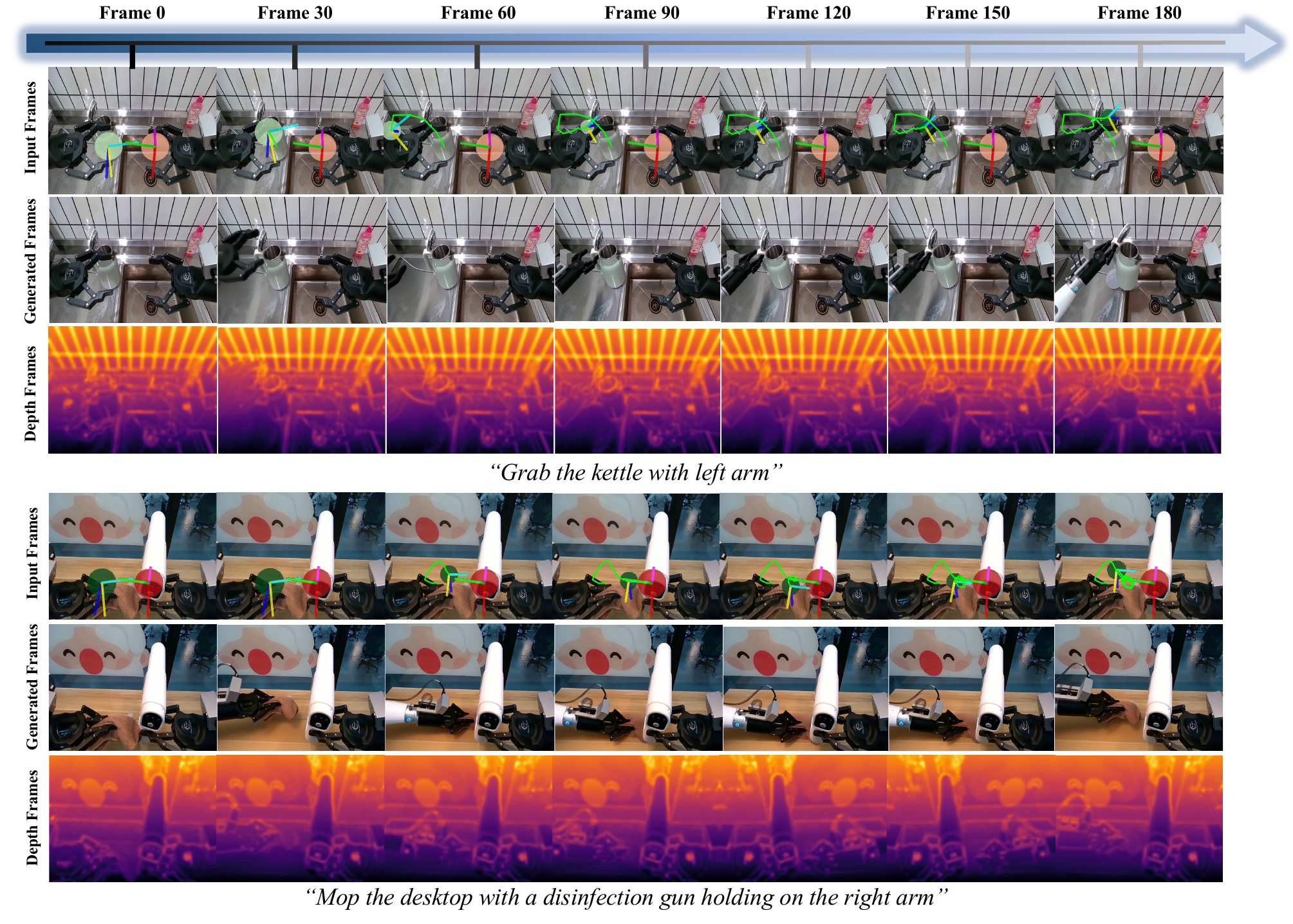}
	\caption{Qualitative results on the AgiBot-Challenge2026 benchmark. For diverse manipulation tasks, \methodname{} rolls out future frames conditioned on the executed robot actions. The generated end-effector and object motions closely track the ground-truth trajectories, and the predicted frames remain sharp and temporally coherent across the rollout.}
	\label{fig:agibot-visual}
\end{figure*}

\begin{figure*}[!t]
	\centering
	\includegraphics[width=\textwidth]{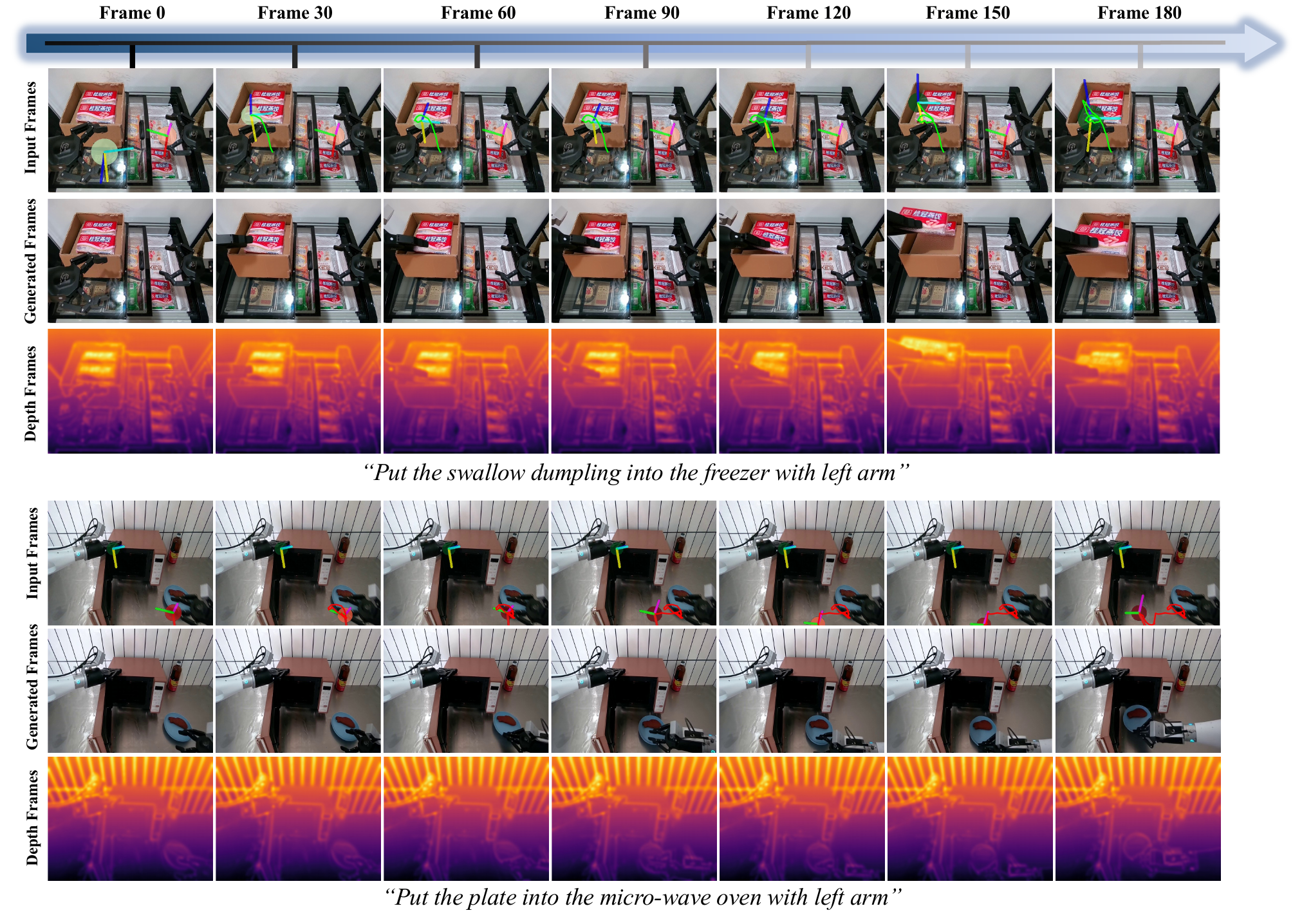}
	\caption{Additional qualitative results on the AgiBot-Challenge2026 benchmark. Across further manipulation tasks, \methodname{} produces action-faithful rollouts in which object and gripper motions follow the commanded actions while preserving fine visual detail over time.}
	\label{fig:agibot-visual2}
\end{figure*}

On the AgiBot-Challenge2026 leaderboard, \methodname{} achieves an EWMScore of 0.8245, ranking second overall and surpassing established teams such as Wild Path and VIPL-GENUN. Notably, it attains the best Scene Consistency score (0.9041), showing that our explicit cross-view geometric reasoning translates directly into superior multi-view coherence under action conditioning. The nDTW score of 0.9531 indicates that the generated trajectories closely track the ground-truth action sequences, validating our action-map conditioning. While NeoVerse-ABot edges ahead on EWMScore and PSNR, \methodname{} overtakes it on Scene Consistency (+0.67\%), the metric most directly tied to 3D consistency, which is our primary design objective. \figref{fig:agibot-visual} shows representative action-conditioned rollouts, where the generated observations faithfully reflect the commanded robot actions and maintain physically plausible scene dynamics over time.

\subsection{Text-Conditioned Multi-View Generation}
\label{sec:exp-text}

We evaluate text-conditioned generation on the AgiBot-World benchmark, comparing against three state-of-the-art baselines: Genie-Envisioner~\cite{liao2025genieenvisioner}, Cosmos-Predict2.5~\cite{nvidia2025cosmospredict25}, and Wan2.1~\cite{wan2025wan}. Results are summarized in \tabref{tab:text-cond}.

\begin{table*}[!t]
	\centering
	\caption{Text-conditioned multi-view generation results on the AgiBot-World benchmark. Best results are in \textbf{bold}, second best are \underline{underlined}. $\uparrow$ indicates higher is better, $\downarrow$ indicates lower is better.}
	\label{tab:text-cond}
	\renewcommand{\arraystretch}{1.15}
	\resizebox{\textwidth}{!}{%
		\begin{tabular}{l c c c c c c c}
			\toprule
			Method & SSIM $\uparrow$ & LPIPS $\downarrow$ & FID $\downarrow$ & FVD $\downarrow$ & Scen Cons. $\uparrow$ & Geometric $\uparrow$ & MEt3R $\downarrow$ \\
			\midrule
			Genie-Envisioner~\cite{liao2025genieenvisioner} & \underline{0.7445} & 0.3345 & 83.7847 & 207.2025 & \textbf{0.9231} & 0.5327 & \underline{15.75} \\
			Cosmos-Predict2~\cite{nvidia2025cosmos} & 0.5870 & \underline{0.3251} & 58.2837 & 188.6350 & 0.8456 & 0.4824 & 17.47 \\
			Wan2.1~\cite{wan2025wan} & 0.5715 & 0.3354 & \underline{56.4735} & \underline{184.2186} & 0.8617 & \underline{0.4716} & 16.59 \\
			\textbf{\methodname{} (Ours)} & \textbf{0.7683} & \textbf{0.1844} & \textbf{45.0389} & \textbf{175.7778} & \underline{0.9041} & \textbf{0.4056} & \textbf{14.20} \\
			\bottomrule
		\end{tabular}%
	}
\end{table*}

\paragraph{Evaluation Metrics.}
We report seven metrics spanning perceptual quality, distributional fidelity, and 3D geometric consistency:
\begin{itemize}[nosep]
	\item \textbf{SSIM}: Structural similarity measuring pixel-level correspondence.
	\item \textbf{LPIPS}: Learned perceptual similarity in deep feature space (lower is better).
	\item \textbf{FID}: Fr\'echet Inception Distance measuring frame-level distributional quality.
	\item \textbf{FVD}: Fr\'echet Video Distance capturing temporal distributional coherence.
	\item \textbf{Scene Consistency}: Temporal semantic coherence measured by DINOv2 feature similarity.
	\item \textbf{Geometric}: Temporal geometric error measured by Sampson epipolar distance (lower is better).
	\item \textbf{MEt3R}: 3D consistency via point cloud cross-projection (lower is better).
\end{itemize}

\begin{figure*}[!t]
	\centering
	\includegraphics[width=\textwidth]{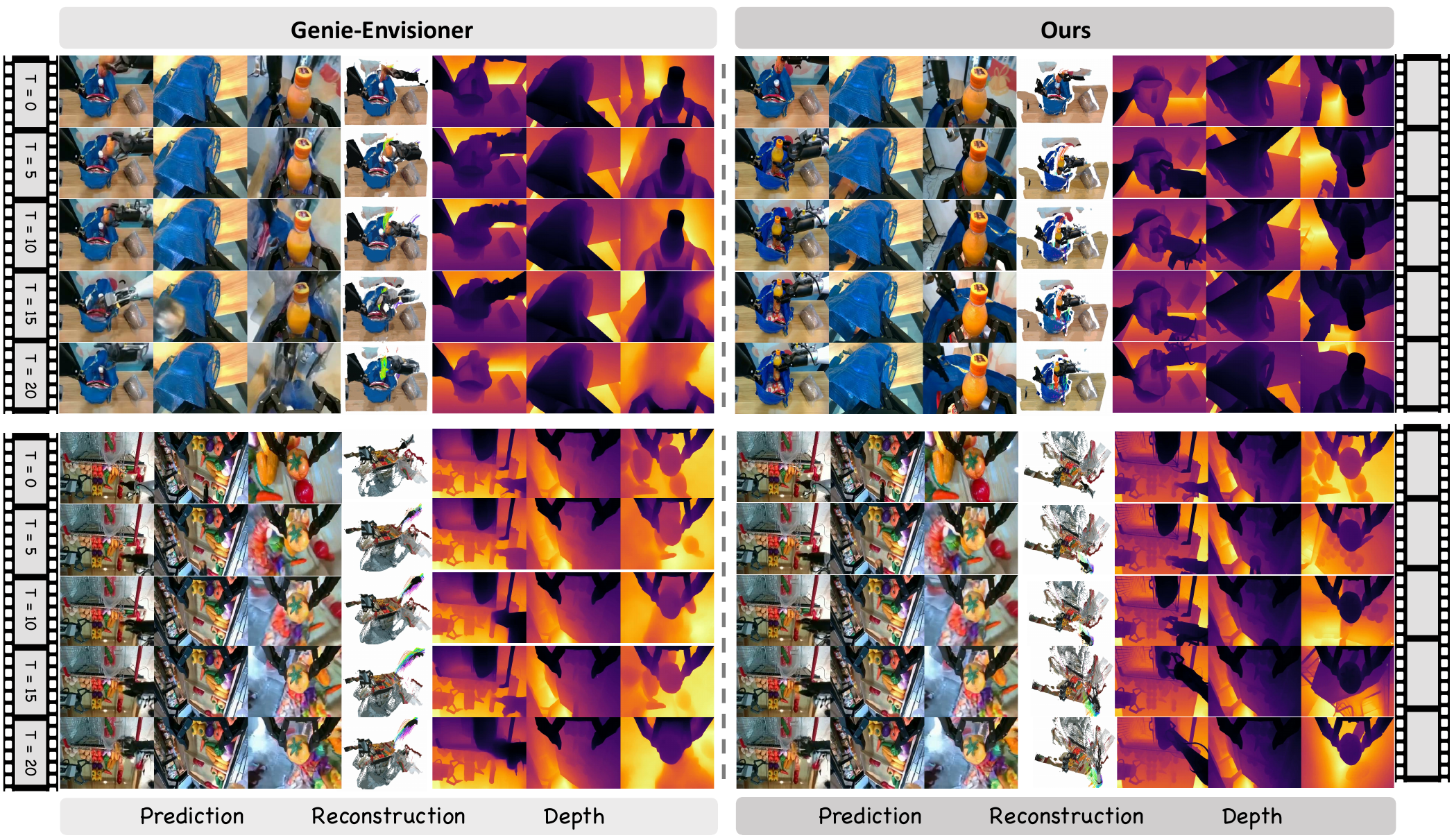}
	\caption{Qualitative comparison of multi-view video generation. For each scene, we show generated frames from two viewpoints. Compared to Genie-Envisioner, \methodname{} produces geometrically consistent cross-view outputs with coherent object placement, depth structure, and texture alignment across viewpoints.}
	\label{fig:visual-comparison}
\end{figure*}

As shown in \figref{fig:visual-comparison}, \methodname{} produces multi-view frames with markedly stronger cross-view geometric consistency. Where the baselines exhibit visible object drift and texture misalignment between viewpoints, \methodname{} preserves coherent 3D structure across views.

\methodname{} attains the best score on 6 of the 7 metrics. On perceptual quality, it reaches an SSIM of 0.7683 and an LPIPS of 0.1844, surpassing the second-best method (Genie-Envisioner) by 3.2\% in SSIM and 45\% in LPIPS, evidence of substantially sharper and more structurally faithful frames. On distributional fidelity, \methodname{} achieves an FID of 45.04, a 20\% improvement over Wan2.1 (56.47), indicating that its generations closely match the real data distribution.

Most tellingly, \methodname{} obtains the best MEt3R score of 14.20, a metric that directly quantifies multi-view 3D reconstruction error. This 10\% improvement over the second-best Genie-Envisioner (15.75) confirms that our three components (Geo-RoPE, Geometry-Aware Cross-View Attention, and Latent 3D-REPA) jointly inject geometric consistency into the generation process. The Geometric consistency score of 0.4056 (where lower denotes better cross-view alignment) corroborates this advantage. Genie-Envisioner attains the highest Scene Consistency score (0.9231), owing to its explicit text-grounding mechanism, while \methodname{} remains competitive at 0.9041, a marginal gap on scene consistency in exchange for a decisive lead on every geometric measure.

\subsection{Ablation Study}
\label{sec:ablation}

Our central claim is that multi-view 3D consistency requires two remedies acting at complementary levels: an architectural pathway for inter-view communication (Geometry-Aware Cross-View Attention, shaped by Geo-RoPE) and a training objective that enforces 3D-consistent content (Latent 3D-REPA), and that \emph{neither alone is sufficient}. To test this directly, we ablate the two components on the AgiBot-World benchmark, starting from the backbone with flat multi-view token concatenation and adding each remedy in isolation and in combination. Results are reported in \tabref{tab:ablation}.

\begin{table}[!t]
	\centering
	\small
	\caption{Ablation of the architectural pathway (Geometry-Aware Cross-View Attention with Geo-RoPE, ``CVA'') and the geometric objective (Latent 3D-REPA, ``REPA'') on the AgiBot-World benchmark. $\Delta$ denotes MEt3R improvement over the backbone. Best results are in \textbf{bold}.}
	\label{tab:ablation}
	\renewcommand{\arraystretch}{1.2}
	\begin{tabular}{cc cccc c}
		\toprule
		CVA & REPA & SSIM $\uparrow$ & LPIPS $\downarrow$ & FID $\downarrow$ & MEt3R $\downarrow$ & $\Delta$ \\
		\midrule
		\xmark & \xmark & 0.6912 & 0.2783 & 53.17 & 16.84 & --- \\
		\cmark & \xmark & 0.7204 & 0.2361 & 50.02 & 15.91 & 0.93 \\
		\xmark & \cmark & 0.7156 & 0.2447 & 49.88 & 16.12 & 0.72 \\
		\cmark & \cmark & \textbf{0.7683} & \textbf{0.1844} & \textbf{45.04} & \textbf{14.20} & \textbf{2.64} \\
		\bottomrule
	\end{tabular}
\end{table}

The results support our claim along three lines. \emph{First, each remedy alone yields only a modest gain.} Adding the communication pathway without geometric supervision (row 2) improves MEt3R by 0.93, but the pathway, lacking an explicit 3D objective, partially settles into texture-copying shortcuts that limit its benefit. Adding the geometric objective without an inter-view pathway (row 3) improves MEt3R by 0.72, but the per-view geometric signal has no route to propagate across viewpoints, so cross-view inconsistencies persist. \emph{Second, the combination is super-additive.} The full model improves MEt3R by 2.64, substantially exceeding the sum of the individual gains ($0.93 + 0.72 = 1.65$). This non-additive jump is the empirical signature of the reinforcing loop analyzed in \secref{sec:synergy}: the pathway transmits information while the objective makes that information 3D-consistent, and only their coupling enforces consistency that propagates coherently across all views. \emph{Third, perceptual quality tracks the same pattern}: SSIM, LPIPS, and FID all improve most when both components are present, confirming that the geometric gains do not come at the expense of visual fidelity.

\section{Conclusion}
\label{sec:conclusion}

We presented \methodname{}, a framework for achieving 3D-consistent multi-view generation in world foundation models for robotic manipulation. Our analysis establishes that multi-view 3D consistency requires two remedies acting at complementary levels: an architectural pathway for inter-view communication and a training-objective signal that enforces 3D-consistent content. Geometry-Aware Cross-View Attention, shaped by Geometric Rotary Position Embedding, opens the pathway through which viewpoints exchange information along geometrically corresponding tokens; Latent 3D-REPA supplies the geometric learning signal that makes the exchanged content faithful to true 3D structure. We showed that neither remedy alone suffices: a pathway without geometric supervision degenerates into trivial shortcuts, while a geometric prior without an inter-view pathway cannot propagate constraints across views. Their combination, addressing both the architectural and objective levels at once, yields consistent multi-view 3D generation.

Built upon the DiT-based world foundation model, \methodname{} attains state-of-the-art multi-view 3D consistency on robotic manipulation benchmarks, leading on reconstruction-based, geometric, and scene-consistency metrics across both text- and action-conditioned settings. This improved consistency carries over to downstream embodied tasks: model-based planning benefits from physically plausible imagined trajectories, world action models exhibit stronger action-visual causal alignment, and multi-view policy post-training yields more effective manipulation policies.

Several promising directions remain for future work. First, incorporating \emph{physical interaction modeling}, such as contact dynamics, deformable objects, and fluid simulation, would push 3D consistency beyond geometry into physics-aware world modeling. Second, scaling to \emph{long-horizon planning} scenarios will require maintaining 3D consistency over extended temporal rollouts, potentially through hierarchical or recurrent architectures. Third, by coupling our world model with a World Action Model (WAM), we aim to build a \emph{world-model-driven data closed loop} for embodied intelligence: the world model generates diverse imagined experiences, the WAM learns from these experiences to improve its policy, and the improved policy in turn collects higher-quality real-world data to further refine the world model, enabling continuous self-improvement and autonomous evolution of embodied agents. Fourth, we plan to develop \emph{industrial manufacturing foundation models} built upon our world modeling framework, targeting applications such as dynamic scheduling of production lines and real-time control of manufacturing processes, where accurate physical simulation and multi-view monitoring are critical for intelligent manufacturing.

\section*{Contributions}

\textbf{Core Contributors:} Yuhang Huang, Jiazhao Zhang, Xuan Lv, Junyan Xu, Zhiyuan Yu, Ruizhen Hu, Kai Xu

\textbf{Contributors:} Wancheng Feng, Shilong Zou, Hewen Xiao, Ziqiao Zhou, Kaiyun Huang, Zhiyu Peng, Juzhan Xu, Hang Zhao, Zhibin Zhu, Chenyang Zhu, Renjiao Yi, Yifei Huang, Douhui Wu, Yan Zhang, Kexu Cheng, Chunhe Song, Yunzhi Xue, Xiuhong Zhang, Leitao Guo, Yunji Chen, Bin Wu, Haibin Yu

\textbf{Corresponding Author:} Kai Xu

\bibliographystyle{unsrtnat}
\bibliography{references}

\begin{thebibliography}{60}
\providecommand{\natexlab}[1]{#1}
\providecommand{\url}[1]{\texttt{#1}}
\expandafter\ifx\csname urlstyle\endcsname\relax
  \providecommand{\doi}[1]{doi: #1}\else
  \providecommand{\doi}{doi: \begingroup \urlstyle{rm}\Url}\fi

\bibitem[Ha and Schmidhuber(2018)]{ha2018world}
David Ha and J{\"u}rgen Schmidhuber.
\newblock World models.
\newblock \emph{arXiv preprint arXiv:1803.10122}, 2018.

\bibitem[Hafner et~al.(2020)Hafner, Lillicrap, Ba, and Norouzi]{hafner2020dreamer}
Danijar Hafner, Timothy Lillicrap, Jimmy Ba, and Mohammad Norouzi.
\newblock Dream to control: Learning behaviors by latent imagination.
\newblock In \emph{International Conference on Learning Representations (ICLR)}, 2020.

\bibitem[NVIDIA et~al.(2025)NVIDIA, Agarwal, Ali, Bala, Balaji, Barker, Cai, Chattopadhyay, Chen, Cui, et~al.]{nvidia2025cosmos}
NVIDIA, Niket Agarwal, Arslan Ali, Maciej Bala, Yogesh Balaji, Erik Barker, Tiffany Cai, Prithvijit Chattopadhyay, Yongxin Chen, Yin Cui, et~al.
\newblock {Cosmos} world foundation model platform for physical {AI}.
\newblock \emph{arXiv preprint arXiv:2501.03575}, 2025.

\bibitem[NVIDIA(2026)]{nvidia2026cosmos3}
NVIDIA.
\newblock {Cosmos 3}: Omnimodal world models for physical {AI}.
\newblock \emph{arXiv preprint arXiv:2606.02800}, 2026.

\bibitem[Liu et~al.(2024{\natexlab{a}})Liu, Zhang, Li, Yan, Gao, Chen, Yuan, Huang, Sun, Gao, He, and Sun]{liu2024sora}
Yixin Liu, Kai Zhang, Yuan Li, Zhiling Yan, Chujie Gao, Ruoxi Chen, Zhengqing Yuan, Yue Huang, Hanchi Sun, Jianfeng Gao, Lifang He, and Lichao Sun.
\newblock {Sora}: A review on background, technology, limitations, and opportunities of large vision models.
\newblock \emph{arXiv preprint arXiv:2402.17177}, 2024{\natexlab{a}}.

\bibitem[Yang et~al.(2025)Yang, Teng, Zheng, Ding, Huang, Xu, Yang, Hong, Zhang, Feng, et~al.]{yang2024cogvideox}
Zhuoyi Yang, Jiayan Teng, Wendi Zheng, Ming Ding, Shiyu Huang, Jiazheng Xu, Yuanming Yang, Wenyi Hong, Xiaohan Zhang, Guanyu Feng, et~al.
\newblock {CogVideoX}: Text-to-video diffusion models with an expert transformer.
\newblock In \emph{International Conference on Learning Representations (ICLR)}, 2025.

\bibitem[Gao et~al.(2024{\natexlab{a}})Gao, Yang, Chen, Chitta, Qiu, Geiger, Zhang, and Li]{gao2024vista}
Shenyuan Gao, Jiazhi Yang, Li~Chen, Kashyap Chitta, Yihang Qiu, Andreas Geiger, Jun Zhang, and Hongyang Li.
\newblock {Vista}: A generalizable driving world model with high fidelity and versatile controllability.
\newblock In \emph{Advances in Neural Information Processing Systems}, volume~37, 2024{\natexlab{a}}.

\bibitem[Kong et~al.(2024)Kong, Tian, Zhang, Min, Dai, Zhou, Xiong, Li, Wu, Zhang, et~al.]{kong2024hunyuanvideo}
Weijie Kong, Qi~Tian, Zijian Zhang, Rox Min, Zuozhuo Dai, Jin Zhou, Jiangfeng Xiong, Xin Li, Bo~Wu, Jianwei Zhang, et~al.
\newblock {HunyuanVideo}: A systematic framework for large video generative models.
\newblock \emph{arXiv preprint arXiv:2412.03603}, 2024.

\bibitem[Wan et~al.(2025)Wan, Wang, Ai, Wen, Mao, Xie, Chen, Yu, Zhao, Yang, et~al.]{wan2025wan}
Team Wan, Ang Wang, Baole Ai, Bin Wen, Chaojie Mao, Chen-Wei Xie, Di~Chen, Feiwu Yu, Haiming Zhao, Jianxiao Yang, et~al.
\newblock Wan: Open and advanced large-scale video generative models.
\newblock \emph{arXiv preprint arXiv:2503.20314}, 2025.

\bibitem[Wu et~al.(2022)Wu, Escontrela, Hafner, Goldberg, and Abbeel]{wu2023daydreamer}
Philipp Wu, Alejandro Escontrela, Danijar Hafner, Ken Goldberg, and Pieter Abbeel.
\newblock {DayDreamer}: World models for physical robot learning.
\newblock In \emph{Proceedings of the Conference on Robot Learning (CoRL)}, volume 205, pages 2226--2240, 2022.

\bibitem[Du et~al.(2023)Du, Yang, Dai, Dai, Nachum, Tenenbaum, Schuurmans, and Abbeel]{du2024learning}
Yilun Du, Sherry Yang, Bo~Dai, Hanjun Dai, Ofir Nachum, Josh Tenenbaum, Dale Schuurmans, and Pieter Abbeel.
\newblock Learning universal policies via text-guided video generation.
\newblock In \emph{Advances in Neural Information Processing Systems}, volume~36, 2023.

\bibitem[Zhou et~al.(2024)Zhou, Du, Chen, Li, Yeung, and Gan]{zhou2024robodreamer}
Siyuan Zhou, Yilun Du, Jiaben Chen, Yandong Li, Dit-Yan Yeung, and Chuang Gan.
\newblock {RoboDreamer}: Learning compositional world models for robot imagination.
\newblock In \emph{Proceedings of the 41st International Conference on Machine Learning (ICML)}, 2024.

\bibitem[Huang et~al.(2025{\natexlab{a}})Huang, Zhang, Zou, Liu, Hu, and Xu]{huang2025ladiwm}
Yuhang Huang, Jiazhao Zhang, Shilong Zou, Xinwang Liu, Ruizhen Hu, and Kai Xu.
\newblock {LaDi-WM}: A latent diffusion-based world model for predictive manipulation.
\newblock \emph{arXiv preprint arXiv:2505.11528}, 2025{\natexlab{a}}.

\bibitem[Peebles and Xie(2023)]{peebles2023dit}
William Peebles and Saining Xie.
\newblock Scalable diffusion models with transformers.
\newblock In \emph{Proceedings of the IEEE/CVF International Conference on Computer Vision (ICCV)}, pages 4195--4205, 2023.

\bibitem[Zhang et~al.(2026)Zhang, Chen, Liu, Ding, Xu, Zou, Liao, Hu, Ren, Zhang, et~al.]{zhang2026pelican}
Yi~Zhang, Yinda Chen, Che Liu, Zeyuan Ding, Jin Xu, Shilong Zou, Junwei Liao, Jiayu Hu, Xiancong Ren, Xiaopeng Zhang, et~al.
\newblock {Pelican-Unified 1.0}: A unified embodied intelligence model for understanding, reasoning, imagination and action.
\newblock \emph{arXiv preprint arXiv:2605.15153}, 2026.

\bibitem[Brohan et~al.(2023{\natexlab{a}})Brohan, Brown, Carbajal, Chebotar, Dabis, Finn, Gopalakrishnan, Hausman, Herzog, Hsu, et~al.]{brohan2023rt1}
Anthony Brohan, Noah Brown, Justice Carbajal, Yevgen Chebotar, Joseph Dabis, Chelsea Finn, Keerthana Gopalakrishnan, Karol Hausman, Alex Herzog, Jasmine Hsu, et~al.
\newblock {RT-1}: Robotics transformer for real-world control at scale.
\newblock In \emph{Robotics: Science and Systems (RSS)}, 2023{\natexlab{a}}.

\bibitem[Brohan et~al.(2023{\natexlab{b}})Brohan, Brown, Carbajal, Chebotar, Chen, Choromanski, Ding, Driess, Dubey, Finn, et~al.]{brohan2023rt2}
Anthony Brohan, Noah Brown, Justice Carbajal, Yevgen Chebotar, Xi~Chen, Krzysztof Choromanski, Tianli Ding, Danny Driess, Avinava Dubey, Chelsea Finn, et~al.
\newblock {RT-2}: Vision-language-action models transfer web knowledge to robotic control.
\newblock In \emph{Proceedings of the Conference on Robot Learning (CoRL)}, volume 229, pages 2165--2183, 2023{\natexlab{b}}.

\bibitem[{Open X-Embodiment Collaboration}(2024)]{collaboration2023openx}
{Open X-Embodiment Collaboration}.
\newblock Open {X}-embodiment: Robotic learning datasets and {RT-X} models.
\newblock In \emph{Proceedings of the IEEE International Conference on Robotics and Automation (ICRA)}, 2024.

\bibitem[Chi et~al.(2023)Chi, Feng, Du, Xu, Cousineau, Burchfiel, and Song]{chi2023diffusionpolicy}
Cheng Chi, Siyuan Feng, Yilun Du, Zhenjia Xu, Eric Cousineau, Benjamin Burchfiel, and Shuran Song.
\newblock Diffusion policy: Visuomotor policy learning via action diffusion.
\newblock In \emph{Robotics: Science and Systems (RSS)}, 2023.

\bibitem[Bruce et~al.(2024)Bruce, Dennis, Edwards, Parker-Holder, Shi, Hughes, Lai, Mavalankar, Steigerwald, Apps, et~al.]{bruce2024genie}
Jake Bruce, Michael Dennis, Ashley Edwards, Jack Parker-Holder, Yuge Shi, Edward Hughes, Matthew Lai, Aditi Mavalankar, Richie Steigerwald, Chris Apps, et~al.
\newblock {Genie}: Generative interactive environments.
\newblock In \emph{Proceedings of the 41st International Conference on Machine Learning}, 2024.

\bibitem[Wu et~al.(2024{\natexlab{a}})Wu, Yin, Feng, He, Li, Hao, and Long]{wu2024ivideogpt}
Jialong Wu, Shaofeng Yin, Ningya Feng, Xu~He, Dong Li, Jianye Hao, and Mingsheng Long.
\newblock {iVideoGPT}: Interactive {VideoGPT}s are scalable world models.
\newblock In \emph{Advances in Neural Information Processing Systems}, volume~37, 2024{\natexlab{a}}.

\bibitem[Su et~al.(2024)Su, Lu, Pan, Wen, and Liu]{su2024rope}
Jianlin Su, Yu~Lu, Shengfeng Pan, Bo~Wen, and Yunfeng Liu.
\newblock {RoFormer}: Enhanced transformer with rotary position embedding.
\newblock \emph{Neurocomputing}, 568:\penalty0 127063, 2024.

\bibitem[He et~al.(2025)He, Xu, Guo, Wetzstein, Dai, Li, and Yang]{he2024cameractrl}
Hao He, Yinghao Xu, Yuwei Guo, Gordon Wetzstein, Bo~Dai, Hongsheng Li, and Ceyuan Yang.
\newblock {CameraCtrl}: Enabling camera control for video diffusion models.
\newblock In \emph{International Conference on Learning Representations (ICLR)}, 2025.

\bibitem[Yu et~al.(2025{\natexlab{a}})Yu, Kwak, Jang, Jeong, Huang, Shin, and Xie]{yu2024repa}
Sihyun Yu, Sangkyung Kwak, Huiwon Jang, Jongheon Jeong, Jonathan Huang, Jinwoo Shin, and Saining Xie.
\newblock {REPA}: Representation alignment for generation: Training diffusion transformers is easier than you think.
\newblock In \emph{International Conference on Learning Representations}, 2025{\natexlab{a}}.

\bibitem[Blattmann et~al.(2023)Blattmann, Dockhorn, Kulal, Mendelevitch, Kilian, Lorenz, Levi, English, Voleti, Letts, et~al.]{blattmann2023svd}
Andreas Blattmann, Tim Dockhorn, Sumith Kulal, Daniel Mendelevitch, Maciej Kilian, Dominik Lorenz, Yam Levi, Zion English, Vikram Voleti, Adam Letts, et~al.
\newblock Stable video diffusion: Scaling latent video diffusion models to large datasets.
\newblock \emph{arXiv preprint arXiv:2311.15127}, 2023.

\bibitem[Alonso et~al.(2024)Alonso, Jelley, Micheli, Kanervisto, Storkey, Pearce, and Fleuret]{alonso2024diamond}
Eloi Alonso, Adam Jelley, Vincent Micheli, Anssi Kanervisto, Amos Storkey, Tim Pearce, and Francois Fleuret.
\newblock {DIAMOND}: Diffusion for world modeling: Visual details matter in atari.
\newblock In \emph{Advances in Neural Information Processing Systems}, volume~37, 2024.

\bibitem[Yang et~al.(2024{\natexlab{a}})Yang, Du, Ghasemipour, Tompson, Kaelbling, Schuurmans, and Abbeel]{yang2023unisim}
Sherry Yang, Yilun Du, Kamyar Ghasemipour, Jonathan Tompson, Leslie Kaelbling, Dale Schuurmans, and Pieter Abbeel.
\newblock Learning interactive real-world simulators.
\newblock In \emph{International Conference on Learning Representations (ICLR)}, 2024{\natexlab{a}}.

\bibitem[Parker-Holder et~al.(2024)Parker-Holder, Spencer, Ball, Bruce, Dasagi, Holsheimer, Kaplanis, Moufarek, Scully, Shar, et~al.]{parkerholder2024genie2}
Jack Parker-Holder, Stephen Spencer, Philip Ball, Jake Bruce, Vibhavari Dasagi, Kristian Holsheimer, Christos Kaplanis, Alexandre Moufarek, Guy Scully, Jeremy Shar, et~al.
\newblock {Genie 2}: A large-scale foundation world model.
\newblock Google DeepMind Blog, 2024.
\newblock \url{https://deepmind.google/blog/genie-2-a-large-scale-foundation-world-model/}.

\bibitem[Cheang et~al.(2024)Cheang, Chen, Jing, Kong, Li, Li, Liu, Wu, Xu, Yang, Zhang, and Zhu]{cheang2024gr2}
Chi-Lam Cheang, Guangzeng Chen, Ya~Jing, Tao Kong, Hang Li, Yifeng Li, Yuxiao Liu, Hongtao Wu, Jiafeng Xu, Yichu Yang, Hanbo Zhang, and Minzhao Zhu.
\newblock {GR-2}: A generative video-language-action model with web-scale knowledge for robot manipulation.
\newblock \emph{arXiv preprint arXiv:2410.06158}, 2024.

\bibitem[Zhu et~al.(2025)Zhu, Wu, Guo, et~al.]{zhu2025irasim}
Fangqi Zhu, Hongtao Wu, Song Guo, et~al.
\newblock {IRASim}: Learning interactive real-robot action simulators.
\newblock In \emph{Proceedings of the IEEE/CVF International Conference on Computer Vision (ICCV)}, 2025.

\bibitem[Huang et~al.(2025{\natexlab{b}})Huang, Chen, Zhou, Chen, Jiang, Hu, Liao, Gao, Li, Yao, and Ren]{huang2025enerverse}
Siyuan Huang, Liliang Chen, Pengfei Zhou, Shengcong Chen, Zhengkai Jiang, Yue Hu, Yue Liao, Peng Gao, Hongsheng Li, Maoqing Yao, and Guanghui Ren.
\newblock {EnerVerse}: Envisioning embodied future space for robotics manipulation.
\newblock In \emph{Advances in Neural Information Processing Systems}, 2025{\natexlab{b}}.

\bibitem[Bu et~al.(2025)Bu, Cai, Chen, Cui, Ding, Feng, Gao, He, Hu, Huang, et~al.]{bu2025agibotworld}
Qingwen Bu, Jisong Cai, Li~Chen, Xiuqi Cui, Yan Ding, Siyuan Feng, Shenyuan Gao, Xindong He, Xuan Hu, Xu~Huang, et~al.
\newblock {AgiBot World Colosseo}: A large-scale manipulation platform for scalable and intelligent embodied systems.
\newblock In \emph{Proceedings of the IEEE/RSJ International Conference on Intelligent Robots and Systems (IROS)}, 2025.

\bibitem[Shang et~al.(2026)Shang, Li, Ma, Su, Jin, Wang, Jin, Zhang, Tang, Su, et~al.]{shang2026worldarena}
Yu~Shang, Zhuohang Li, Yiding Ma, Weikang Su, Xin Jin, Ziyou Wang, Lei Jin, Xin Zhang, Yinzhou Tang, Haisheng Su, et~al.
\newblock {WorldArena}: A unified benchmark for evaluating perception and functional utility of embodied world models.
\newblock \emph{arXiv preprint arXiv:2602.08971}, 2026.

\bibitem[Liu et~al.(2023)Liu, Wu, Van~Hoorick, Tokmakov, Zakharov, and Vondrick]{liu2023zero123}
Ruoshi Liu, Rundi Wu, Basile Van~Hoorick, Pavel Tokmakov, Sergey Zakharov, and Carl Vondrick.
\newblock {Zero-1-to-3}: Zero-shot one image to {3D} object.
\newblock In \emph{Proceedings of the IEEE/CVF International Conference on Computer Vision (ICCV)}, pages 9298--9309, 2023.

\bibitem[Liu et~al.(2024{\natexlab{b}})Liu, Lin, Zeng, Long, Liu, Komura, and Wang]{liu2023syncdreamer}
Yuan Liu, Cheng Lin, Zijiao Zeng, Xiaoxiao Long, Lingjie Liu, Taku Komura, and Wenping Wang.
\newblock {SyncDreamer}: Generating multiview-consistent images from a single-view image.
\newblock In \emph{International Conference on Learning Representations (ICLR)}, 2024{\natexlab{b}}.

\bibitem[Shi et~al.(2024)Shi, Wang, Ye, Mai, Li, and Yang]{shi2023mvdream}
Yichun Shi, Peng Wang, Jianglong Ye, Long Mai, Kejie Li, and Xiao Yang.
\newblock {MVDream}: Multi-view diffusion for {3D} generation.
\newblock In \emph{International Conference on Learning Representations}, 2024.

\bibitem[Tang et~al.(2023)Tang, Zhang, Chen, Wang, and Furukawa]{tang2023mvdiffusion}
Shitao Tang, Fuyang Zhang, Jiacheng Chen, Peng Wang, and Yasutaka Furukawa.
\newblock {MVDiffusion}: Enabling holistic multi-view image generation with correspondence-aware diffusion.
\newblock In \emph{Advances in Neural Information Processing Systems}, volume~36, 2023.

\bibitem[Voleti et~al.(2024)Voleti, Yao, Boss, Letts, Pankratz, Tochilkin, Laforte, Rombach, and Jampani]{voleti2024sv3d}
Vikram Voleti, Chun-Han Yao, Mark Boss, Adam Letts, David Pankratz, Dmitrii Tochilkin, Christian Laforte, Robin Rombach, and Varun Jampani.
\newblock {SV3D}: Novel multi-view synthesis and {3D} generation from a single image using latent video diffusion.
\newblock In \emph{Proceedings of the European Conference on Computer Vision (ECCV)}, 2024.

\bibitem[Xie et~al.(2024)Xie, Yao, Voleti, Jiang, and Jampani]{xie2024sv4d}
Yiming Xie, Chun-Han Yao, Vikram Voleti, Huaizu Jiang, and Varun Jampani.
\newblock {SV4D}: Dynamic {3D} content generation with multi-frame and multi-view consistency.
\newblock \emph{arXiv preprint arXiv:2407.17470}, 2024.

\bibitem[Gao et~al.(2024{\natexlab{b}})Gao, Holynski, Henzler, Brussee, Martin-Brualla, Srinivasan, Barron, and Poole]{gao2024cat3d}
Ruiqi Gao, Aleksander Holynski, Philipp Henzler, Arthur Brussee, Ricardo Martin-Brualla, Pratul Srinivasan, Jonathan~T. Barron, and Ben Poole.
\newblock {CAT3D}: Create anything in {3D} with multi-view diffusion models.
\newblock In \emph{Advances in Neural Information Processing Systems}, volume~37, 2024{\natexlab{b}}.

\bibitem[Yu et~al.(2025{\natexlab{b}})Yu, Xing, Yuan, Hu, Li, Huang, Gao, Wong, Shan, and Tian]{yu2024viewcrafter}
Wangbo Yu, Jinbo Xing, Li~Yuan, Wenbo Hu, Xiaoyu Li, Zhipeng Huang, Xiangjun Gao, Tien-Tsin Wong, Ying Shan, and Yonghong Tian.
\newblock {ViewCrafter}: Taming video diffusion models for high-fidelity novel view synthesis.
\newblock \emph{IEEE Transactions on Pattern Analysis and Machine Intelligence}, 2025{\natexlab{b}}.

\bibitem[Mildenhall et~al.(2020)Mildenhall, Srinivasan, Tancik, Barron, Ramamoorthi, and Ng]{mildenhall2020nerf}
Ben Mildenhall, Pratul~P. Srinivasan, Matthew Tancik, Jonathan~T. Barron, Ravi Ramamoorthi, and Ren Ng.
\newblock {NeRF}: Representing scenes as neural radiance fields for view synthesis.
\newblock In \emph{Proceedings of the European Conference on Computer Vision (ECCV)}, pages 405--421, 2020.

\bibitem[Kerbl et~al.(2023)Kerbl, Kopanas, Leimk{\"u}hler, and Drettakis]{kerbl2023gaussiansplatting}
Bernhard Kerbl, Georgios Kopanas, Thomas Leimk{\"u}hler, and George Drettakis.
\newblock {3D} gaussian splatting for real-time radiance field rendering.
\newblock \emph{ACM Transactions on Graphics}, 42\penalty0 (4):\penalty0 1--14, 2023.

\bibitem[Wang et~al.(2024)Wang, Leroy, Cabon, Chidlovskii, and Revaud]{wang2024dust3r}
Shuzhe Wang, Vincent Leroy, Yohann Cabon, Boris Chidlovskii, and Jerome Revaud.
\newblock {DUSt3R}: Geometric {3D} vision made easy.
\newblock In \emph{Proceedings of the IEEE/CVF Conference on Computer Vision and Pattern Recognition (CVPR)}, pages 20697--20709, 2024.

\bibitem[Leroy et~al.(2024)Leroy, Cabon, and Revaud]{leroy2024mast3r}
Vincent Leroy, Yohann Cabon, and Jerome Revaud.
\newblock Grounding image matching in {3D} with {MASt3R}.
\newblock In \emph{Proceedings of the European Conference on Computer Vision (ECCV)}, 2024.

\bibitem[Yang et~al.(2024{\natexlab{b}})Yang, Kang, Huang, Xu, Feng, and Zhao]{yang2024depthanything}
Lihe Yang, Bingyi Kang, Zilong Huang, Xiaogang Xu, Jiashi Feng, and Hengshuang Zhao.
\newblock {Depth Anything}: Unleashing the power of large-scale unlabeled data.
\newblock In \emph{Proceedings of the IEEE/CVF Conference on Computer Vision and Pattern Recognition (CVPR)}, pages 10371--10381, 2024{\natexlab{b}}.

\bibitem[Yang et~al.(2024{\natexlab{c}})Yang, Kang, Huang, Zhao, Xu, Feng, and Zhao]{yang2024depthanythingv2}
Lihe Yang, Bingyi Kang, Zilong Huang, Zhen Zhao, Xiaogang Xu, Jiashi Feng, and Hengshuang Zhao.
\newblock {Depth Anything V2}.
\newblock In \emph{Advances in Neural Information Processing Systems}, volume~37, 2024{\natexlab{c}}.

\bibitem[Lin et~al.(2026)Lin, Chen, Liew, Chen, Li, Shi, Feng, and Kang]{lin2025depthanything3}
Haotong Lin, Sili Chen, Jun~Hao Liew, Donny~Y. Chen, Zhenyu Li, Guang Shi, Jiashi Feng, and Bingyi Kang.
\newblock {Depth Anything 3}: Recovering the visual space from any views.
\newblock In \emph{International Conference on Learning Representations (ICLR)}, 2026.

\bibitem[Wang et~al.(2025)Wang, Chen, Karaev, Vedaldi, Rupprecht, and Novotny]{wang2025vggt}
Jianyuan Wang, Minghao Chen, Nikita Karaev, Andrea Vedaldi, Christian Rupprecht, and David Novotny.
\newblock {VGGT}: Visual geometry grounded transformer.
\newblock In \emph{Proceedings of the IEEE/CVF Conference on Computer Vision and Pattern Recognition (CVPR)}, 2025.

\bibitem[Lipman et~al.(2023)Lipman, Chen, Ben-Hamu, Nickel, and Le]{lipman2023flow}
Yaron Lipman, Ricky T.~Q. Chen, Heli Ben-Hamu, Maximilian Nickel, and Matt Le.
\newblock Flow matching for generative modeling.
\newblock In \emph{International Conference on Learning Representations (ICLR)}, 2023.

\bibitem[Esser et~al.(2024)Esser, Kulal, Blattmann, Entezari, M{\"u}ller, Saini, Levi, Lorenz, Sauer, Boesel, et~al.]{esser2024sd3}
Patrick Esser, Sumith Kulal, Andreas Blattmann, Rahim Entezari, Jonas M{\"u}ller, Harry Saini, Yam Levi, Dominik Lorenz, Axel Sauer, Frederic Boesel, et~al.
\newblock Scaling rectified flow transformers for high-resolution image synthesis.
\newblock In \emph{Proceedings of the 41st International Conference on Machine Learning (ICML)}, 2024.

\bibitem[Jiang et~al.(2025{\natexlab{a}})Jiang, Chen, Huang, Chen, Zhou, Liao, He, Liu, Li, Yao, and Ren]{huang2025evac}
Yuxin Jiang, Shengcong Chen, Siyuan Huang, Liliang Chen, Pengfei Zhou, Yue Liao, Xindong He, Chiming Liu, Hongsheng Li, Maoqing Yao, and Guanghui Ren.
\newblock {EnerVerse-AC}: Envisioning embodied environments with action condition.
\newblock \emph{arXiv preprint arXiv:2505.09723}, 2025{\natexlab{a}}.

\bibitem[NVIDIA(2025{\natexlab{a}})]{nvidia2025cosmospredict25}
NVIDIA.
\newblock World simulation with video foundation models for physical {AI}.
\newblock \emph{arXiv preprint arXiv:2511.00062}, 2025{\natexlab{a}}.

\bibitem[NVIDIA(2025{\natexlab{b}})]{nvidia2025cosmosreason1}
NVIDIA.
\newblock {Cosmos-Reason1}: From physical common sense to embodied reasoning.
\newblock \emph{arXiv preprint arXiv:2503.15558}, 2025{\natexlab{b}}.

\bibitem[Wu et~al.(2024{\natexlab{b}})Wu, Hou, Liu, Che, Ju, Yang, Li, Zhao, Xu, Yang, et~al.]{wu2024robomind}
Kun Wu, Chengkai Hou, Jiaming Liu, Zhengping Che, Xiaozhu Ju, Zhuqin Yang, Meng Li, Yinuo Zhao, Zhiyuan Xu, Guang Yang, et~al.
\newblock {RoboMIND}: Benchmark on multi-embodiment intelligence normative data for robot manipulation.
\newblock \emph{arXiv preprint arXiv:2412.13877}, 2024{\natexlab{b}}.

\bibitem[Jiang et~al.(2025{\natexlab{b}})Jiang, Yuan, Liu, Lu, Cui, Liu, Cheng, Gao, Xu, and Zhao]{jiang2025galaxea}
Tao Jiang, Tianyuan Yuan, Yicheng Liu, Chenhao Lu, Jianning Cui, Xiao Liu, Shuiqi Cheng, Jiyang Gao, Huazhe Xu, and Hang Zhao.
\newblock Galaxea open-world dataset and {G0} dual-system {VLA} model.
\newblock \emph{arXiv preprint arXiv:2509.00576}, 2025{\natexlab{b}}.

\bibitem[Mu et~al.(2025)Mu, Chen, Gao, Lan, Zou, Lin, Xie, and Luo]{mu2025robotwin}
Yao Mu, Tianxing Chen, Zeyu Gao, Zhiqian Lan, Yude Zou, Lunkai Lin, Zhiqiang Xie, and Ping Luo.
\newblock {RoboTwin}: Dual-arm robot benchmark with generative digital twins.
\newblock In \emph{Proceedings of the IEEE/CVF Conference on Computer Vision and Pattern Recognition (CVPR)}, 2025.

\bibitem[Wu et~al.(2025)Wu, Liu, Xie, Wang, Li, et~al.]{wu2025robocoin}
Shihan Wu, Xuecheng Liu, Shaoxuan Xie, Pengwei Wang, Xinghang Li, et~al.
\newblock {RoboCOIN}: An open-sourced bimanual robotic data collection for integrated manipulation.
\newblock \emph{arXiv preprint arXiv:2511.17441}, 2025.

\bibitem[Team et~al.(2026)Team, Xiang, Bao, Liu, Tan, Bi, Li, Liu, Pang, Jing, Liu, Cai, Cui, Zhao, Wang, Huang, Feng, Rong, Wang, and Zhu]{motubrainteam2026motubrainadvancedworldaction}
MotuBrain Team, Chendong Xiang, Fan Bao, Haitian Liu, Hengkai Tan, Hongzhe Bi, James Li, Jiabao Liu, Jingrui Pang, Kiro Jing, Louis Liu, Mengchen Cai, Rongxu Cui, Ruowen Zhao, Runqing Wang, Shuhe Huang, Yao Feng, Yinze Rong, Zeyuan Wang, and Jun Zhu.
\newblock Motubrain: An advanced world action model for robot control, 2026.
\newblock URL \url{https://arxiv.org/abs/2604.27792}.

\bibitem[Liao et~al.(2025)Liao, Jiang, Chen, Huang, Zhou, Chen, Liu, He, Liu, Yao, Ren, and Li]{liao2025genieenvisioner}
Yue Liao, Yuxin Jiang, Liliang Chen, Siyuan Huang, Pengfei Zhou, Shengcong Chen, Chiming Liu, Xindong He, Yi~Liu, Maoqing Yao, Guanghui Ren, and Hongsheng Li.
\newblock {Genie Envisioner}: A unified world foundation platform for robotic manipulation.
\newblock \emph{arXiv preprint arXiv:2508.05635}, 2025.

\end{thebibliography}

\end{document}